\documentclass[pdflatex,sn-vancouver-ay]{sn-jnl}

\usepackage{amsmath,amssymb,amsfonts}
\usepackage{algorithmic}
\usepackage{graphicx}
\usepackage{textcomp}
\usepackage{makecell}
\usepackage{cuted}
\usepackage{mathtools}
\usepackage{booktabs}
\usepackage{xcolor}
\usepackage{enumitem}
\usepackage{placeins}
\usepackage{url}
\usepackage{hyperref}
\usepackage{caption}
\usepackage{silence}

\usepackage{subcaption}

\newcommand{\code}[1]{{\ttfamily #1}}
\DeclareMathOperator{\NoisyOR}{Noisy-OR}

\newcommand{\pr}[1]{#1}
\newcommand{\rebut}[1]{#1}

\theoremstyle{thmstyleone}%
\theoremstyle{thmstyletwo}%

\theoremstyle{thmstylethree}%

\begin{document}

\title[SimSUM]{SimSUM -- Simulated Benchmark with Structured and Unstructured Medical Records}


\author*[1]{\fnm{Paloma} \sur{Rabaey}}\email{paloma.rabaey@ugent.be}

\author[2]{\fnm{Stefan} \sur{Heytens}}\email{stefan.heytens@ugent.be}

\author[1]{\fnm{Thomas} \sur{Demeester}}\email{thomas.demeester@ugent.be}

\affil[1]{\orgdiv{Dept. of Information Technology}, \orgname{Ghent University -- imec}, \orgaddress{ \city{Ghent}, \country{Belgium}}}

\affil[2]{\orgdiv{Dept. of Public Health and Primary Care}, \orgname{Ghent University}, \orgaddress{\city{Ghent}, \country{Belgium}}}


\abstract{
\textbf{Background:} Clinical information extraction, which involves structuring clinical concepts from unstructured medical text, remains a challenging problem that could benefit from the inclusion of tabular background information available in electronic health records. Existing open-source datasets lack explicit links between structured features and clinical concepts in the text, motivating the need for a new research dataset.\\
\textbf{Methods:} We introduce SimSUM, a benchmark dataset of 10,000 simulated patient records that link unstructured clinical notes with structured background variables. Each record simulates a patient encounter in the domain of respiratory diseases and includes tabular data (e.g., symptoms, diagnoses, underlying conditions) generated from a Bayesian network whose structure and parameters are defined by domain experts. A large language model (GPT-4o) is prompted to generate a clinical note describing the encounter, including symptoms and relevant context. These notes are annotated with span-level symptom mentions. We conduct an expert evaluation to assess note quality and run baseline predictive models on both the tabular and textual data.\\
\textbf{Conclusion:} The SimSUM dataset is primarily designed to support research on clinical information extraction in the presence of tabular background variables, which can be linked through domain knowledge to concepts of interest to be extracted from the text—namely, symptoms in the case of SimSUM. Secondary uses include research on the automation of clinical reasoning over both tabular data and text, causal effect estimation in the presence of tabular and/or textual confounders, and multi-modal synthetic data generation. SimSUM is not intended for training clinical decision support systems or production-grade models, but rather to facilitate reproducible research in a simplified and controlled setting. The dataset is available at \url{https://github.com/prabaey/SimSUM}.
}

\keywords{Bayesian network, clinical notes, clinical information extraction, electronic health records, large language model, tabular data, simulated benchmark.}



\maketitle

\section{Background}
\label{sec:introduction}

\begin{figure*}
\centering
\includegraphics[width=\textwidth]{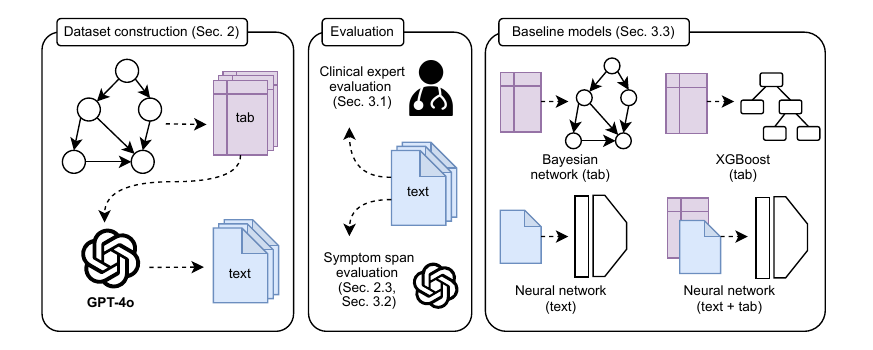}
\caption{\rebut{General overview of the structure of this paper. We first describe the construction of SimSUM, a simulated dataset combining structured tabular data and unstructured clinical text (Section \ref{sec:methodology}). We then evaluate the clinical notes through both a clinical expert review (Section \ref{sec:expert_evaluation}) and automated span-level symptom extraction (Sections \ref{sec:span_extraction} and \ref{sec:utility_spans}). Finally, we present four baseline predictive models that take as input the tabular data, the textual data, or both (Section \ref{sec:CIE_baselines}).}}
\label{fig:overview}
\end{figure*}

\begin{figure*}
\centering
\includegraphics[width=0.9\textwidth]{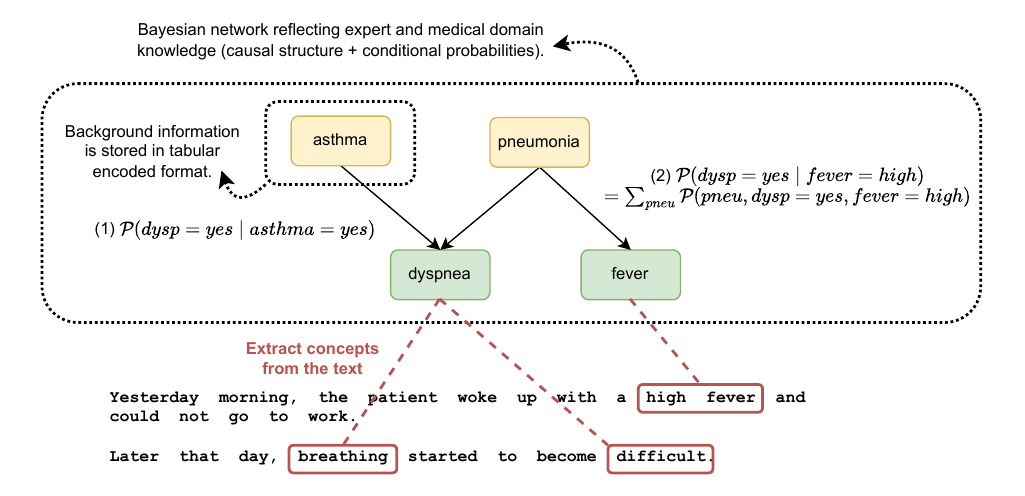}
\caption{
We have a clinical description of a patient encounter from which we want to extract some concepts, in this case the symptoms experienced by the patient. Some symptoms might be easy to extract using text-matching, like ``high fever''. Other symptoms are not mentioned verbatim and are therefore harder to extract, like dyspnea. In this case, additional information on the patient, present in encoded format in the tabular portion of the EHR, together with domain knowledge, may help. We illustrate this idea with two examples. \\
\textbf{Example 1}: We know that the patient has asthma. Domain knowledge may tell us that the probability of experiencing dyspnea when one has asthma \rebut{(Equation (1))} is 90\%, \rebut{thereby increasing the prior probability of encountering dyspnea in the text. By integrating this knowledge in the information extraction module, it can more accurately predict the posterior probability of encountering dyspnea in the text}.\\
\textbf{Example 2}: We know that the patient is experiencing high fever. Domain knowledge may tell us that a high fever often co-occurs with dyspnea due to their common cause, which is pneumonia. Even if we do not know that the patient has pneumonia, the probability of dyspnea being mentioned in the text increases as a result of observing high fever. By modeling the joint probability of dyspnea, fever and pneumonia using a Bayesian network, we can get the exact probability of $\mathcal{P}(\text{\textit{dyspnea = yes}} \mid \text{\textit{fever = high}})$ by summing over the \rebut{possible} presence \rebut{and absence} of pneumonia \rebut{in a procedure called Bayesian inference} (\rebut{Equation (2)}, \cite{koller2009probabilistic}).}
\label{fig:IE_with_background}
\end{figure*}

\textbf{Electronic health records} (EHRs) are a gold mine of information, containing a mix of structured tabular variables\footnote{\rebut{In the remainder of our work, we use the terms ``structured tabular data'' and ``tabular features'' interchangeably.}} (medication, diagnosis codes, lab results\dots) and free unstructured text (detailed clinical notes from physicians, nurses\dots) \citep{extracting_information_EMR}. These EHRs form a valuable basis for training clinical decision support systems, (partially) automating essential processes in the clinical world, such as diagnosis, writing treatment plans, and more \citep{CDS_infectious_diseases, clinical_text_classification, medBERT, BEHRT, multimodal_ICD}. While large language models can help leverage the potential of the unstructured text portion of the EHR \citep{combining_structured_unstructured, multimodal_matters, clinicalBERT, clinicalT5, MedPalm, BioMistral}, these black box systems lack interpretability \citep{black_box, explainability_LLMs, opportunities_challenges_biomedicine}. In high-risk clinical applications, it can be argued that one should prefer more robust and transparent systems built on simpler, feature-based models, like regression models, decision trees, or Bayesian networks \citep{rudin2019stop, CML_for_healthcare, explainable_trees}. However, such models cannot directly deal with unstructured text and require tabular features as an input. For this reason, automated clinical information extraction (CIE) \citep{extracting_information_EMR, clinical_information_extraction} is an essential tool for building large structured datasets that can serve as training data for such systems. \\

\textbf{Clinical information extraction} remains a challenging task due to the complex nature of clinical notes \citep{medical_information_extraction}. These often leave out important contextual details 
which an automated system would need in order to correctly extract concepts from the text.
\rebut{State-of-the-art clinical information extraction systems are predominantly built on pre-trained language models. Earlier methods relied on models pre-trained for entity and relation extraction in clinical text \citep{pretrained_biomedical}, while more recent approaches draw on large language models that have absorbed substantial medical knowledge during training \citep{LLMs_information_extraction}. In both paradigms, domain knowledge is acquired implicitly rather than being explicitly encoded. Moreover, current systems typically overlook complementary information available in structured patient records. 
}
We propose that CIE could benefit from leveraging two additional sources of information, apart from the unstructured text itself. On the one hand, a range of tabular features are already encoded in the EHR. These contain information related to a particular patient visit (e.g. partially encoded symptoms or diagnosis codes), as well as information on the medical history of the patient. On the other hand, we can connect this encoded background information with the concepts we are trying to extract from the text, using a Bayesian network (BN) that represents medical domain knowledge. 
\rebut{While ontologies and controlled vocabularies are commonly used to capture standardized concepts and semantic relationships \citep{medical_informed_ML, medical_KG}, they lack the ability to model causal dependencies or uncertainty. An expert-defined Bayesian network (BN) complements these resources by modeling how clinical factors influence one another and with what probabilistic strength \citep{BN_healthcare}.} 
A visual example that helps illustrate this idea is shown in Fig. \ref{fig:IE_with_background}.\\

To investigate and implement this idea, we need a \textbf{clinical dataset which (i) contains a mix of tabular data and unstructured text, where (ii) the tabular data and the concepts we aim to extract from the text can be linked through domain knowledge.} While open-source datasets like MIMIC-III \citep{mimic3} and MIMIC-IV \citep{mimic4} contain this mix, they are not a perfect fit. First, the area of intensive care in which the data was collected is very extensive, making it hard to isolate a specific use-case for which the domain knowledge could be listed. Second, the portion of the dataset which is encoded into tabular features is often driven by billing needs \citep{mimic4}, rather than completeness or accuracy, and does not contain any encoded symptoms, which are concepts that could be interesting to extract from the text for application in clinical decision support systems. Third, the link between the tabular features and the concepts mentioned in the text might be inconsistent due to system design or human errors \citep{EHRcon, improving_ehr}. Finally, the EHRs in MIMIC \rebut{are time series, and while temporal information is a standard aspect of EHRs, it adds significant modeling complexity, making simpler datasets preferable for prototyping novel methods.} Other existing datasets linking unstructured clinical text to structured features include BioDEX \citep{biodex}, a large set of papers describing adverse drug events, as well as TCGA-Reports \citep{tcga_reports}, a set of cancer pathology reports, both accompanied by tabular patient descriptors and extracted biological features. However, in both cases, it is not trivial to devise a BN representing the relevant expert knowledge, partially because that knowledge is not fully understood yet. \\

In this work, we simulate a sufficiently realistic dataset that addresses some of these shortcomings, enabling research on incorporating domain knowledge for improved CIE in the presence of tabular variables. Our dataset, called \textbf{SimSUM\footnote{An earlier version of this dataset was published under the name SynSUM \citep{synsum_old}. It has since been renamed to SimSUM to avoid confusion with synthetic data generated from real data, and to emphasize the simulated nature of the dataset.} (Simulated Structured and Unstructured Medical records)} is a self-contained set of artificial EHRs in a primary care setting, fulfilling the following requirements: 
\setlist{nolistsep}
\begin{itemize}[noitemsep]
    \item Each \rebut{record in SimSUM} is an \rebut{extract} representing a single patient encounter, eliminating the time aspect for simplicity. 
    \item Each \rebut{record is made up of} structured tabular data and unstructured text. 
    \item 
    By design, clinical concepts expressed in the text and encoded in the tabular portion of SimSUM are connected through a Bayesian network representing domain knowledge. In this case, the domain is respiratory diseases, with their associated symptoms and underlying conditions.
    \item The unstructured text in SimSUM contains additional context on some of the encoded tabular variables. \rebut{For example, symptoms are stored as structured tabular data, while the text elaborates on the nature of these symptoms, including their severity, onset, and other details.}
\end{itemize}


SimSUM is constructed in the domain of respiratory diseases, simulating patient visits to a primary care doctor. 
We mimic the scenario where the doctor notes down the patient's symptoms in a clinical note, along with some additional context, and stores this in the EHR together with the encoded diagnosis, as well as the encoded symptoms. Additionally, 
the EHR stores tabular background information on the underlying health conditions of the patient.\\ 

It is important to emphasize that \textbf{the SimSUM dataset is not intended for training clinical decision support systems or deploying predictive models in real-world healthcare}. Instead, it provides a controlled research environment for developing and prototyping methods. Although designed to be domain-relevant, SimSUM is entirely synthetic: tabular data is generated via an expert-defined Bayesian network, and clinical notes are produced by a large language model. \pr{It simulates a specific use-case in a limited domain -- respiratory diseases in primary care -- without modeling any temporal dynamics.} SimSUM therefore does not reflect the full variability or complexity of real EHRs, and should not be used to evaluate clinical model performance or train production models. Its value lies in enabling research on integrating background information into clinical information extraction, within a simplified setting with known and controllable ground-truth structure.\\

The remainder of this work is structured \rebut{as shown in Figure \ref{fig:overview}}. \rebut{In short,} Section \ref{sec:methodology} describes how we generated SimSUM by simulating 10{,}000 patient encounters by sampling from an expert-defined BN and prompting a large language model. Section \ref{sec:utility_discussion} analyses the utility of our dataset and discusses its intended use. Finally, Section \ref{sec:discussion} states our main conclusions and highlights SimSUM's limitations. 

\section{Construction and content} \label{sec:methodology}

\begin{figure*}
\centering
\includegraphics[width=0.8\textwidth]{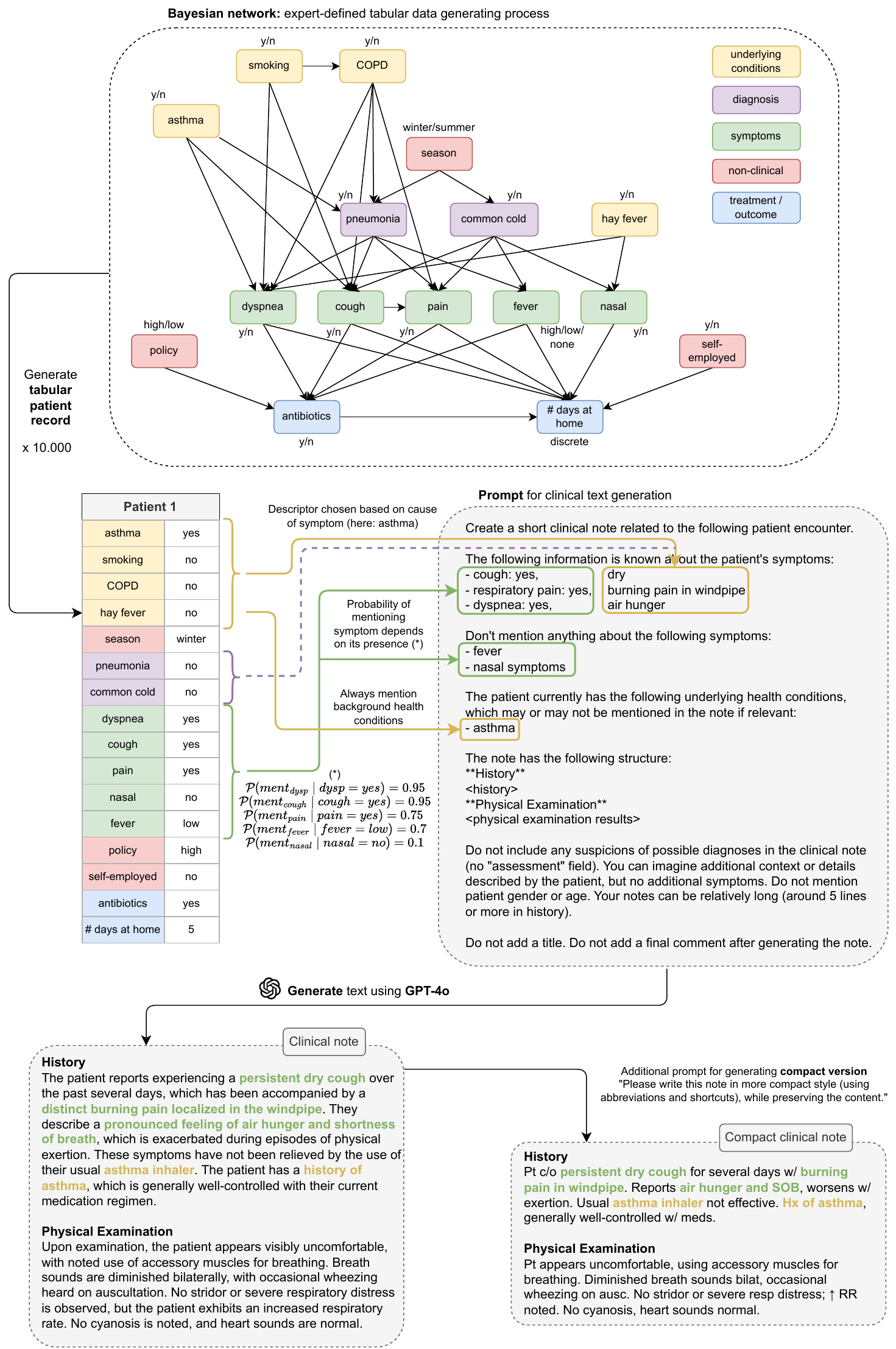}
\caption{Overview of the full data generating process for the SimSUM dataset. First, the tabular portion of the artificial patient record is sampled from a Bayesian network, where both the structure and the conditional probability distributions were defined by an expert. Afterwards, we construct a prompt describing the symptoms experienced by the patient, as well as their underlying health conditions (but no diagnoses). We ask the GPT-4o large language model to generate a clinical note describing this patient encounter. Finally, we ask to generate a more challenging compact version of the note, mimicking the complexity of real clinical notes by prompting the use of abbreviations and shortcuts. We generate $10{,}000$ of these artificial patient records in total.}
\label{fig:data_generation}
\end{figure*}

Our general methodology for generating the artificial patient records is shown in Fig. \ref{fig:data_generation}. 
We now zoom in on the two major parts of this data generating process. First, Section \ref{sec:bayesian_network} describes how we generated the structured tabular variables through an expert-defined Bayesian network. 
Then, Section \ref{sec:prompting_strategy} dives into how the clinical notes were generated by the large language model. Finally, Section \ref{sec:span_extraction} describes how we enhanced our dataset by automatically annotating each note with all spans mentioning each symptom. 

\subsection{Modeling structured tabular variables with a Bayesian network} \label{sec:bayesian_network}

\rebut{For this part of the process, we enlisted an expert with over 30 years of experience in primary care. The expert has a background in developing and applying Bayesian networks in clinical settings and received support from the authors for any technical questions that arose.}\\

\textbf{Causal structure}\quad We asked the expert to define a Directed Acyclic Graph (DAG) which (partially) models the domain of respiratory diseases in primary care, shown in Fig. \ref{fig:data_generation}. In this DAG, a directed arrow between two variables models a causal relation between them. Central to the model are the diagnoses of \textit{pneumonia} and \textit{common\:cold}, which may give rise to five symptoms (\textit{dyspnea}, \textit{cough}, \textit{pain}, \textit{fever} and \textit{nasal}). The expert also modeled some relevant underlying conditions which may render a patient more predisposed to certain diagnoses or symptoms: \textit{asthma}, \textit{smoking}, \textit{COPD} and \textit{hay\:fever}. Based on the symptoms experienced by a patient, a primary care doctor decides whether to prescribe \textit{antibiotics} or not. The presence and severity of the symptoms, as well as the prescription of antibiotics as a treatment, influence the outcome, which is the total number of days that the patient eventually stays home as a result of illness (\textit{days\:at\:home}). Finally, there are some non-clinical variables which exert an external influence on the diagnoses, treatment and outcome (\textit{season}, \textit{policy} and \textit{self{\text -}employed}). Table \ref{tab:variables_description} summarizes all variables and their meaning, as well as their possible values. \rebut{While this model is a simplification of a real-world clinical process, we believe it is sufficiently realistic to fulfill our purpose of generating a simulated dataset.\\} 

\begin{table*}[t]
\centering
\caption{Description of all 16 tabular variables in our dataset, including their type (underlying condition, diagnosis, symptom, non-clinical, treatment or outcome), and their possible values.}
\resizebox{\textwidth}{!}{
\begin{tabular}{llll}
    \toprule
    \textbf{Name} & \textbf{Type} & \textbf{Description} & \textbf{Values} \\
    \midrule
    Asthma & underlying condition & \makecell[lt]{Chronic lung disease in which the airways narrow and swell.} & yes/no \\
    Smoking & underlying condition & \makecell[lt]{Whether the patient is a regular smoker of tobacco.} & yes/no \\
    COPD & underlying condition & \makecell[lt]{Chronic Obstructive Pulmonary Disease, where airflow from the lungs \\is obstructed.} & yes/no \\
    Hay fever & underlying condition & \makecell[lt]{Allergic rhinitis, irritation of the nose caused by an allergen (e.g. pollen).} & yes/no \\
    Season & non-clinical & \makecell[lt]{Season of the year.} & winter/summer \\
    Pneumonia & diagnosis & \makecell[lt]{Infection that inflames the air sacs in one or both lungs.} & yes/no \\
    Common cold & diagnosis & \makecell[lt]{Upper respiratory tract infection, irritation and swelling of the upper \\ airways.} & yes/no \\
    Dyspnea & symptom & \makecell[lt]{Shortness of breath, the feeling of not getting enough air.} & yes/no \\
    Cough & symptom & \makecell[lt]{Any type of cough.
    } & yes/no \\
    Pain & symptom & \makecell[lt]{Pain related to the airways or chest area.} & yes/no \\
    Fever & symptom & \makecell[lt]{Elevation of body temperature.} & high/low/none \\
    Nasal & symptom & \makecell[lt]{Nasal symptoms, such as runny nose or sneezing.} & yes/no \\
    Policy & non-clinical & \makecell[lt]{Whether the clinician has higher or lower prior inclination to prescribe \\ antibiotics. Can be influenced by many factors, such as local policy in \\ their general practice, their own caution towards antibiotics or level of \\ experience.} & high/low \\
    Self-employed & non-clinical & \makecell[lt]{Whether the patient is self-employed, rendering them less inclined to \\ stay home from work for longer periods.} & yes/no \\
    Antibiotics & treatment & \makecell[lt]{Whether any type of antibiotics are prescribed to the patient.} & yes/no \\
    Days at home & outcome & \makecell[lt]{How many days the patient ends up staying home as a result of their \\ symptoms and treatment.} & discrete $(0 - ...)$\\
    \bottomrule
    \bottomrule
\end{tabular}}
\label{tab:variables_description}
\end{table*}

\noindent
\textbf{Probability distribution} \quad To define a data generating mechanism from which we can sample simulated patients, we turn the DAG from Fig. \ref{fig:data_generation} into a Bayesian network by defining a joint probability distribution. This distribution, shown in Equation \eqref{eq:joint_distr}, factorizes into $16$ conditional distributions, one for each variable. 

\begin{equation}
\resizebox{0.9\textwidth}{!}{
$
\begin{aligned}
    &\mathcal{P}_{\text{\textit{joint}}}(\text{\textit{asthma, smoking, ..., antibiotics, \pr{days at home}}}) = \\
    &\mathcal{P}(\text{\textit{asthma}}) \mathcal{P}(\text{\textit{smoking}}) \mathcal{P}(\text{\textit{COPD}} \mid \text{\textit{smoking}})
    \mathcal{P}(\text{\textit{hay fever}}) \mathcal{P}(\text{\textit{season}}) 
    \mathcal{P}(\text{\textit{common cold}} \mid \text{\textit{season}}) \\
    &\mathcal{P}(\text{\textit{pneumonia}} \mid \text{\textit{asthma, COPD, season}})
    \mathcal{P}(\text{\textit{dyspnea}} \mid \text{\textit{asthma, smoking, COPD, pneumonia, hay fever}}) \\
    &\mathcal{P}(\text{\textit{cough}} \mid \text{\textit{asthma, smoking, COPD, pneumonia, common cold}}) \mathcal{P}(\text{\textit{nasal}} \mid \text{\textit{common cold, hay fever}}) \\
    &\mathcal{P}(\text{\textit{pain}} \mid \text{\textit{cough, pneumonia, COPD, common cold}})
    \mathcal{P}(\text{\textit{fever}} \mid \text{\textit{pneumonia, common cold}}) \\
    & \mathcal{P}(\text{\textit{policy}}) \mathcal{P}(\text{\textit{self{\text-}employed}}) 
    \mathcal{P}(\text{\textit{antibiotics}} \mid \text{\textit{policy, dyspnea, cough, pain, fever}}) \\
    &\mathcal{P}(\text{\textit{\pr{days at home}}} \mid \text{\textit{antibiotics, dyspnea, cough, pain, fever, nasal, self{\text-}employed}}) \\
\end{aligned}
$
}
\label{eq:joint_distr}
\end{equation}

~\\
We use four different approaches to parameterize these conditional distributions. \rebut{For this process, we again rely on our expert primary care practitioner. Given that SimSUM is intended as a simplified, controlled benchmark rather than a reflection of real EHR variability, we find input from a single domain expert sufficient to guide the design of its probability distribution.}\\ 

\begin{figure*}[t]
\centering
\includegraphics[width=0.7\textwidth]{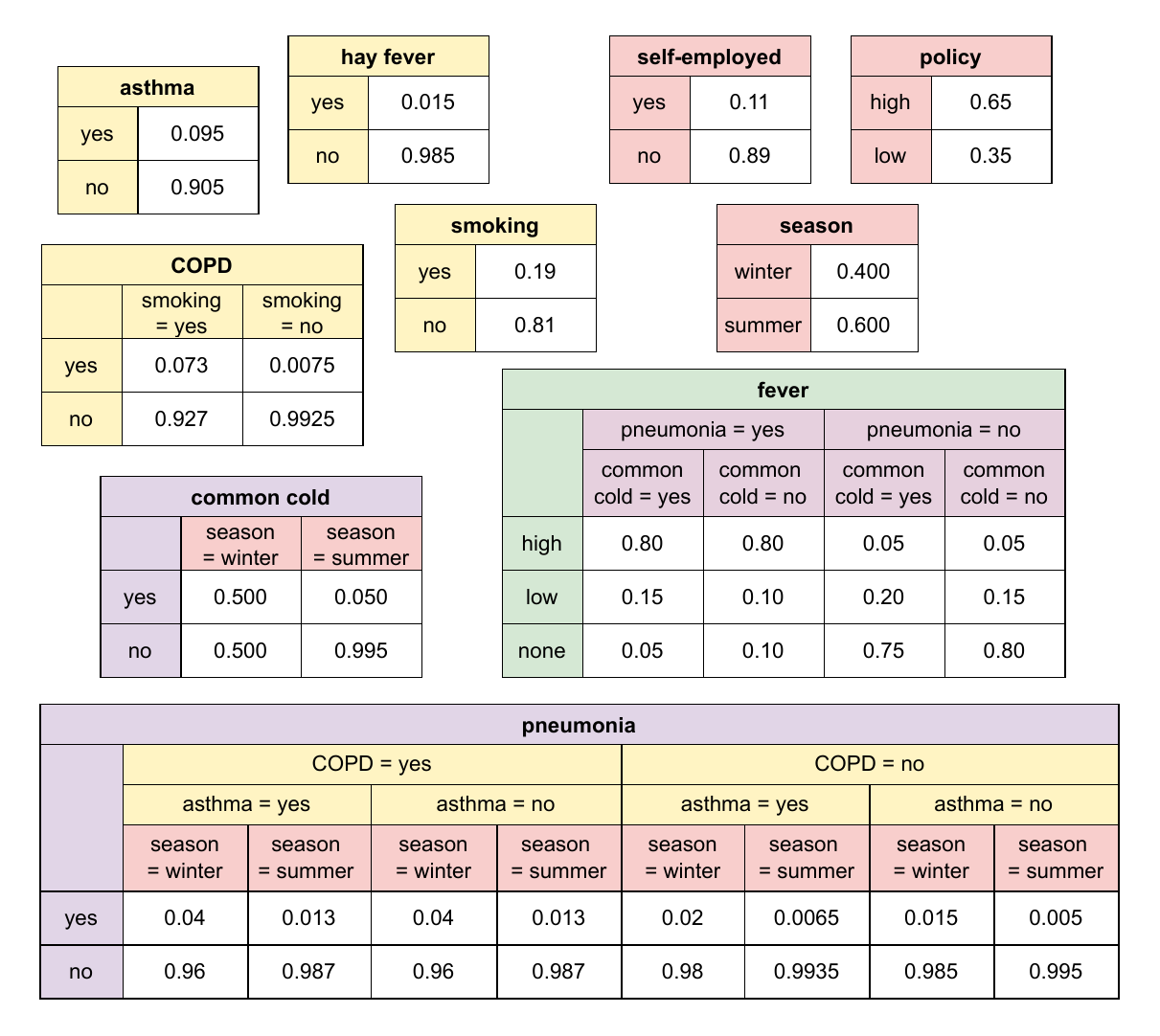}
\caption{Conditional probability tables for the variables \textit{asthma}, \textit{smoking}, \textit{hay\:fever}, \textit{COPD}, \textit{season}, \textit{pneumonia}, \textit{common\:cold}, \textit{fever}, \textit{policy} and \textit{self{\text-}employed}. 
}
\label{fig:CPTs}
\end{figure*}

\noindent
\rebut{\textbf{1. Conditional probability table}\quad When the variable is discrete and has a limited number of parents, we ask the expert to define a conditional probability table (CPT). 
We provide these tables for the variables \textit{asthma}, \textit{smoking}, \textit{hay fever}, \textit{COPD}, \textit{season}, \textit{pneumonia}, \textit{common cold}, \textit{fever}, \textit{policy} and \textit{self-employed} in Figure \ref{fig:CPTs}. The probabilities in the tables were filled in based on the expert's own experience, as well as demographics in their local general practice and Belgium. While we do not expect these probabilities to generalize to the global patient population as a whole, an expert-informed distribution which contains realistic elements suffices for our use-case.}\\

\noindent
\rebut{\textbf{2. Noisy-OR distribution} \quad For categorical variables with many parents, it becomes infeasible to manually fill in the CPT in a clinically meaningful way, because of the large number of possible combinations of parent values. This is the case for the symptoms \textit{dyspnea}, \textit{cough}, \textit{pain} and \textit{nasal} in our BN. To circumvent this problem, we define a Noisy-OR distribution, which assumes an independent causal mechanism behind the activation of a symptom through any of its parents \citep{koller2009probabilistic}. 
This is a reasonable assumption to make in the case of symptoms with multiple possible causes (parents in the BN): a symptom arises in a patient if any of its possible causes succeeds in activating the symptom through its own independent mechanism \citep{noisy_or, koller2009probabilistic}. As shown in Equation \eqref{eq:noisy_OR}, the parameterization of the Noisy-OR distribution rests on choosing each parameter $p_i$, which is the probability that a possible cause $X_i$ activates symptom $Y$. As a special case, $p_0$, also known as the leak probability, is the probability that symptom $Y$ is activated as the result of another unmodeled cause (outside of all $X_i$'s). Note that $x_i$ is $1$ when the cause $X_i$ is present in the patient (``yes'' in our BN), and $0$ if not. }

\begin{equation}
\label{eq:noisy_OR}
\begin{split}
    &\mathcal{P}(Y=1 \mid X_1=x_1, \dots, X_k=x_k) 
    = 1-\mathcal{P}(Y=0 \mid X_1=x_1, \dots., X_k=X_k)\\
    &= 1-(1-p_0)(1-p_1)^{x_1}\dots(1-p_k)^{x_k} 
    =\NoisyOR(p_0, p_1, \dots, p_k; x_1, \dots, x_k) \\
    &\text{with~} x_1, \dots, x_k \in \{0, 1\}
\end{split}
\end{equation}

~\\
\rebut{Equations \eqref{eq:noisy_OR_dysp} through \eqref{eq:noisy_OR_nasal} define such a Noisy-OR distribution for the symptoms \textit{dyspnea}, \textit{cough}, \textit{pain} and \textit{nasal}. Note that the symptom \textit{fever} is fully defined through a CPT, since the expert was able to provide intuition on all possible combinations of its two parent values, eliminating the need for a Noisy-OR distribution. }

\begin{equation}
\begin{split} \label{eq:noisy_OR_dysp}
    &\mathcal{P}(\text{\textit{dysp}} \mid \text{\textit{asthma, smoking, COPD, hayf, pneu}}) 
    = \NoisyOR(p_0=0.05, \\
    &p_{\text{\textit{asthma}}} = 0.9,
    p_{\text{\textit{smoking}}} = 0.3,
    p_{\text{\textit{COPD}}} = 0.9, p_{\text{\textit{hayf}}} = 0.2, p_{\text{\textit{pneu}}} = 0.3)
\end{split}
\end{equation}

\begin{equation}
\begin{split} \label{eq:noisy_OR_cough}
    &\mathcal{P}(\text{\textit{cough}} \mid \text{\textit{asthma, smoking, COPD, pneu, cold}}) = \NoisyOR(p_0=0.07, \\
    &p_{\text{\textit{asthma}}} = 0.3,
    p_{\text{\textit{smoking}}} = 0.6,
    p_{\text{\textit{COPD}}} = 0.4, p_{\text{\textit{pneu}}} = 0.85, p_{\text{\textit{cold}}} = 0.7)
\end{split}
\end{equation}

\begin{equation}
\begin{split} \label{eq:noisy_OR_pain}
    &\mathcal{P}(\text{\textit{pain}} \mid \text{\textit{COPD}}, \text{\textit{cough}}, \text{\textit{pneu}}, \text{\textit{cold}}) 
    = \NoisyOR(p_0=0.05, p_{\text{\textit{COPD}}}
    = 0.15, \\
    &p_{\text{\textit{cough}}} = 0.2,
    p_{\text{\textit{pneu}}} = 0.3, p_{\text{\textit{cold}}} = 0.1)
\end{split}
\end{equation}

\begin{equation}
\begin{split} \label{eq:noisy_OR_nasal}
    &\mathcal{P}(\text{\textit{nasal}} \mid \text{\textit{hayf, cold}})
    = \NoisyOR(p_0=0.1, p_{\text{\textit{hayf}}} = 0.85, p_{\text{\textit{cold}}} = 0.7)
\end{split}
\end{equation}

\noindent
\rebut{\textbf{3. Logistic regression} \quad 
We model the prescription of \textit{antibiotics} by mimicking the way the clinician's suspicion of pneumonia (which needs to be treated with antibiotics) rises when a higher number of symptoms is present in the patient, with some symptoms weighing more than others. Once their level of suspicion reaches a certain threshold, they decide to prescribe treatment. 
As shown in Equation \eqref{eq:antibiotics}, this process can be \rebut{approximated in a simplified way} through a logistic regression model, taking as an input the symptoms \textit{dyspnea}, \textit{cough}, \textit{pain} and \textit{fever}, as well as the variable \textit{policy}. Here, $x_{po}$ (\textit{policy}) can take on the values 1 (high) or 0 (low), $x_d$ (\textit{dyspnea}), $x_c$ (\textit{cough}) and $x_{pa}$ (\textit{pain}) can take on the value 1 (yes) or 0 (no), and $x_f$ (\textit{fever}) can be 2 (high), 1 (low) or 0 (none).}

\begin{equation}\label{eq:antibiotics}
\begin{split} 
    &\mathcal{P}(\text{\textit{antibio}} = \text{\textit{yes}} \mid \text{\textit{policy}}=x_{po}, \text{\textit{dysp}}=x_{d}, \text{\textit{cough}}=x_{c}, 
    \text{\textit{pain}}=x_{pa}, \text{\textit{fever}}=x_{f}) \\
    &= \text{Sigmoid}(-3 + 1 \times x_{po} + 0.8 \times x_d + 0.665 \times x_c
    + 0.665 \times x_{pa} \\
    &+ 0.9 \times (x_f==1) + 2.25 \times (x_f==2)) \\
    &\text{with } x_{po}, x_d, x_c, x_{pa} \in \{0, 1\} \text{, and } x_f \in \{0, 1, 2\}
\end{split}
\end{equation}

\rebut{We decided on the weights in this model with help of the expert. First, we set the bias term 
and the coefficient for \textit{policy} based on the following constraint: if there's no symptoms at all, and \textit{policy} is low, then the probability of prescribing antibiotics (due to some other unmodeled cause) should be around 5\% . If \textit{policy} is high, it should be around 10\%. 
All other coefficients were then chosen by the expert based on the relative importance of the symptoms when deciding to prescribe antibiotics. 
As a final sanity-check, we asked the expert to label a set of test cases with whether or not they would prescribe antibiotics, and compared these with the probability predicted by the model. Table \ref{tab:antibiotics_test_cases} in Appendix \ref{sec:app_BN} shows these results. We see that the predictions made by the model mostly correspond well with the \rebut{desired input-output behavior, indicating} that the proposed coefficients make sense.} \\

\noindent
\rebut{\textbf{4. Poisson regression} \quad In a similar fashion, we model the number of \textit{\pr{days at home}} as a result of the patient's complaints with a Poisson regression model. 
Assuming a non-linear effect of prescribing antibiotics, we define two separate models: one where antibiotics were not prescribed (Equation \eqref{eq:days_home_no_antibiotics}), and one where they were (Equation \eqref{eq:days_home_antibiotics}). Both models take as an input the symptoms \textit{dysp}, \textit{cough}, \textit{pain}, \textit{nasal} and \textit{fever}, as well as the variable \textit{self{\text-}employed}, and predict a mean number of days $\lambda$, which parameterizes the Poisson distribution.}

\begin{equation}\label{eq:days_home_no_antibiotics}
\begin{split} 
    &\mathcal{P}(\text{\textit{\pr{days at home}}} \mid \text{\textit{dysp}}=x_{d}, \text{\textit{cough}}=x_{c}, \text{\textit{pain}}=x_{pa},
    \text{\textit{nasal}}=x_{n}, \text{\textit{fever}}=x_{f}, \\
    &\text{\textit{self{\text-}empl}}=x_{se},\text{\textit{antibio}}=\text{\textit{no}}) 
    = Poisson(\lambda_0) \\
    &\lambda_0 = exp(0.010 + 0.64 \times x_d + 0.35 \times x_c + 0.47 \times x_{pa} + 0.011 \times x_n \\
    &+ 0.81 \times (x_f==1) + 1.23 \times (x_f==2) - 0.5 \times x_{se}) \\
    &\text{with } x_d, x_c, x_{pa}, x_n, x_{se} \in \{0, 1\} \text{, and } x_f \in \{0, 1, 2\}
\end{split}
\end{equation}

\begin{equation} \label{eq:days_home_antibiotics}
\begin{split}
    &\mathcal{P}(\text{\textit{\pr{days at home}}} \mid \text{\textit{dysp}}=x_{d}, \text{\textit{cough}}=x_{c}, \text{\textit{pain}}=x_{pa},
    \text{\textit{nasal}}=x_{n}, \text{\textit{fever}}=x_{f}, \\
    &\text{\textit{selfempl}}=x_{se}, \text{\textit{antibio}}=\text{\textit{yes}}) = Poisson(\lambda_1) \\
    &\lambda_1 = exp(0.16 + 0.51 \times x_d + 0.42 \times x_c + 0.26 \times x_{pa} + 0.0051 \times x_n \\
    &+ 0.24 \times (x_f==1) + 0.57 \times (x_f==2) - 0.5 \times x_{se}) \\
    &\text{with } x_d, x_c, x_{pa}, x_n, x_{se} \in \{0, 1\} \text{, and } x_f \in \{0, 1, 2\}
\end{split}
\end{equation}

\rebut{The coefficients for each model were tuned using gradient descent based on the train cases shown in Table \ref{tab:days_train_cases} in Appendix \ref{sec:app_BN}. The expert was asked to (loosely) label these cases with how long they suspected the patient to stay home as a result of these symptoms. The coefficient for the variable \textit{self{\text-}employed} was tuned manually, based on the assumption that being self-employed would shave some days off the predicted number, regardless of the particular symptoms experienced by the patient. As a sanity check, we compared the mean number of days predicted by the model (parameter $\lambda$ in the Poisson model) with the number of days estimated by the expert for a small set of test cases which were not seen during training. The results are shown in Table \ref{tab:days_test_cases} in Appendix \ref{sec:app_BN}.} \\

\noindent
\textbf{Sampling} \quad 
\pr{Once the joint probability distribution from Equation \eqref{eq:joint_distr} is fully specified,} we can use this Bayesian network to randomly sample the tabular portion of a patient record top-down. Hereby, we start from the root variables without parents at the top and continue further down. Each value is sampled conditionally on the variable's parents' values, using the conditional distributions we have defined. We repeat this process $10{,}000$ times, leaving us with $10{,}000$ artificial patient records consisting of 16 tabular features. 

\subsection{Generating unstructured text with a large language model} \label{sec:prompting_strategy}


Once the tabular patient record has been generated, we prompt a large language model (LLM, in this case GPT-4o) to write a clinical note \pr{based on this fictional patient encounter and its associated tabular variables.} Only the background variables 
and symptoms 
may be directly mentioned in the prompt, while the diagnoses \textit{pneumonia} and \textit{common\:cold} would not yet be known to a clinician who is taking notes during a consultation  
and are therefore left out. Realistically, they can still influence the content of the note through the descriptions of the symptoms that are included in the prompt, as will be explained later. The treatment and outcome are left out of the prompt as well, just like the non-clinical variables, as all of these are typically either unknown or irrelevant at the time of writing the note.\\

\pr{The prompt is generally structured like the example shown in Fig. \ref{fig:data_generation}, and is made up of the following parts. 
Further details and additional example prompts can be found in Appendix \ref{sec:app_prompt_details}.}

\begin{itemize}

\item \pr{First, we list the \textbf{symptoms} which are experienced by the patient. We do not exhaustively list the full set of symptoms, but rather sample the probability that a symptom is mentioned according to a distribution defined by the expert. This renders the notes more realistic, since real notes do not exhaustively mention the presence or absence of all possible symptoms, instead following the narrative of the patient and the subsequent probing of the clinician.}
\item \pr{While we encourage the LLM to invent additional context around the patient's symptoms, we want this context to at least partially relate to the underlying cause of the symptom. For this reason, we include a \textbf{descriptor} next to each symptom, which is randomly sampled from a set of expert-defined phrases that describe the symptom in the case where it results from a particular cause. This way, the diagnoses can influence the content of the note, even when they are not mentioned explicitly in the prompt. The full list of descriptors, as well as further information on how we sample them, is included in Table \ref{tab:symptom_descriptors}.}
\item \pr{The next part of the prompt lists the \textbf{underlying health conditions}, which are assumed to be known up-front, and could therefore influence the content of the note. \item At the end, the prompt lists some \textbf{additional instructions}. We tell the LLM that the note must be structured with a ``History'' portion and a ``Physical examination'' portion. We also ask it not to mention any suspicions of possible diagnoses, nor patient gender or age. Further motivation for these requests can be found in Appendix \ref{sec:app_prompt_details}.\\}
\end{itemize}


\pr{Furthermore, we introduce two additional prompts to extend our dataset and make the notes more realistic.}
\begin{itemize}
\item \pr{Real clinical notes can be challenging, often containing abbreviations, shortcuts and denser sentence structure. To make the notes more challenging, we ask the LLM to create a \textbf{compact version} of each note through an additional prompt, as can be seen in the bottom part of Fig. \ref{fig:data_generation}.} 
\item \pr{We devise an alternative prompting strategy which is used to generate clinical notes in the special case where \textbf{no respiratory symptoms} are present in the patient record. This occurs for around one third of our dataset. In these cases, we encourage the LLM to imagine an alternative reason for the patient visit, unrelated to the respiratory domain. Examples and further details relating to this alternative strategy are included in Appendix \ref{sec:app_prompting_special_cases}.\\}
\end{itemize}

\pr{According to the prompts outlined above, we ask the LLM to generate both a clinical note and its compact version for each of the $10{,}000$ patients in the dataset.} \pr{After generating each note, we use the LLM to perform an additional \textbf{consistency check} to ensure the quality of the generated notes. More specifically, we prompt the LLM once again, showing it the same information on the patient's symptoms and underlying health conditions as in the original prompt. We then ask the LLM to decide whether the generated note is consistent with this information or not. This way, the LLM identified 52 out of $10{,}000$ notes as inconsistent with the prompt, which were regenerated and manually checked.}

\subsection{Automated extraction of symptom spans} \label{sec:span_extraction}

\begin{figure}[t]
\centering
\begin{subfigure}[t]{0.48\linewidth}
    \centering
    \includegraphics[width=\linewidth]{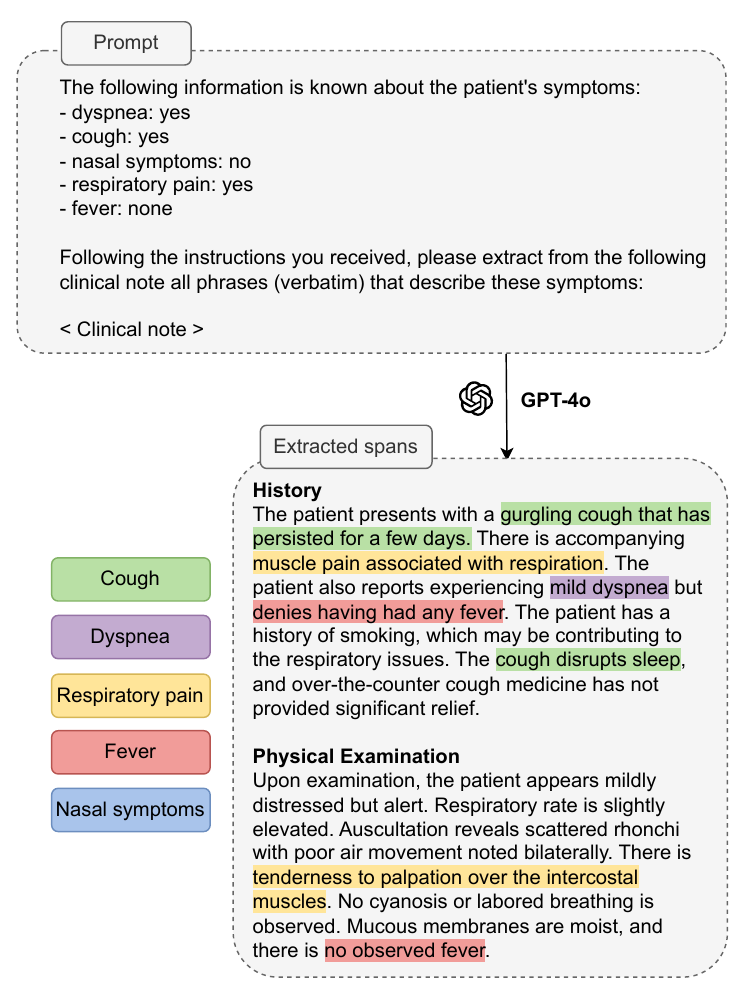}
    \caption{LLM-based annotation of a normal clinical note with symptom spans. The model is prompted to return a JSON object of the form \texttt{\{``symptom'': symptom, ``text'': extracted phrase\}}. For clarity, we highlight the extracted spans in the note rather than displaying the raw JSON. Full task instructions are in Appendix~\ref{app:span_extraction}.}
    \label{fig:span_extraction_normal}
\end{subfigure}
\hfill
\begin{subfigure}[t]{0.48\linewidth}
    \centering
    \includegraphics[width=\linewidth]{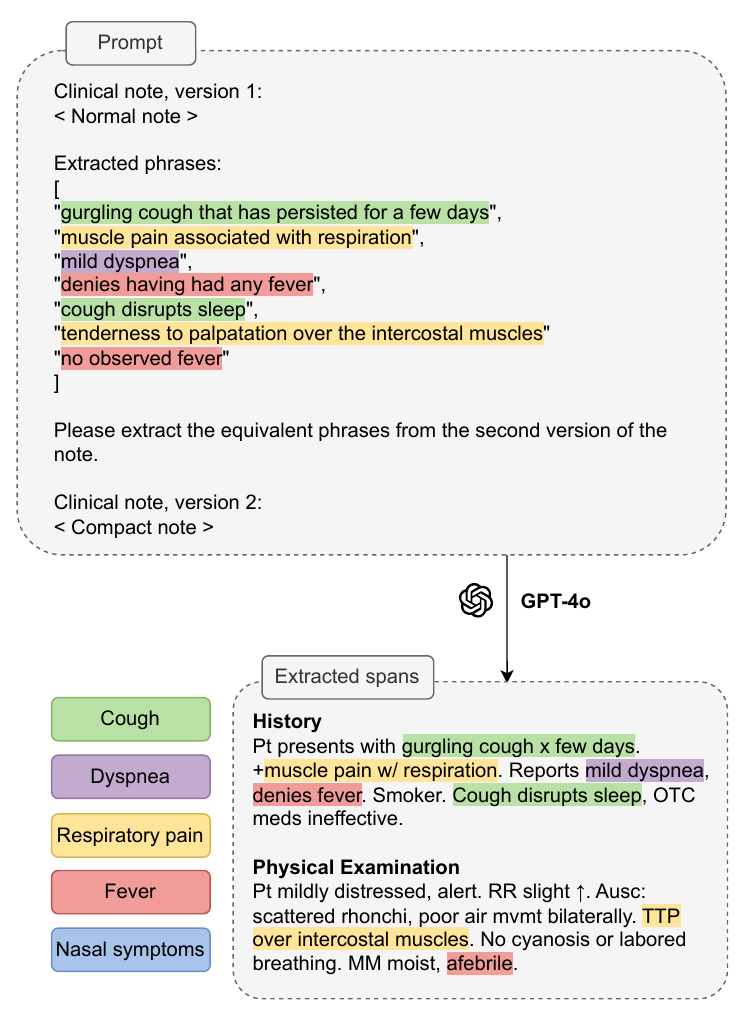}
    \caption{LLM-based annotation of a compact clinical note using spans previously extracted from the corresponding normal note. Full task instructions are provided in Appendix~\ref{app:span_extraction}.}
    \label{fig:span_extraction_compact}
\end{subfigure}
\caption{Prompting strategy for extracting symptom spans from (a) normal and (b) compact clinical notes using a large language model.}
\label{fig:span_extraction_both}
\end{figure}



To further enhance the dataset, we annotated the notes with all spans that mention a symptom. We automatically extracted these phrases from the note using additional LLM calls. \\

\textbf{Normal notes}\quad Fig. \ref{fig:span_extraction_normal} shows the prompt and the extracted phrases for one of the normal notes. We first tell the LLM what symptoms the patient does and does not have, and then ask it to extract phrases that describe these symptoms. We use the system message to send extensive task instructions to the LLM, which we have included in Appendix \ref{app:span_extraction}. We further process the extracted phrases by matching them to the note, disregarding capital letters and punctuation. If no exact match is found, we use another LLM call to correct the mistake, asking it to map the extracted phrase to an exact phrase in the text.  \\

\textbf{Compact notes}\quad For the compact notes, we proceed differently, as shown in Fig. \ref{fig:span_extraction_compact}. We first tell the LLM which phrases were extracted from the normal note, and ask it to match these with the compact version of the note. We found that this worked better than asking it to extract phrases from the compact note from scratch. As with the normal notes, we post-process the extracted phrases by matching them to the text and asking the LLM to correct itself when no exact match is found. \\

\pr{Using the strategy outlined above, we extract all text spans mentioning each of the 5 symptoms from all $10{,}000$ notes in our dataset. In Section \ref{sec:utility_spans}, we use these fine-grained extracted spans to analyze the content of the notes and quantify the presence of each symptom in the textual portion of SimSUM.} 

\section{Utility and discussion}\label{sec:utility_discussion}

We now present a series of experiments to illustrate the utility and potential of our dataset. First, Section \ref{sec:expert_evaluation} reports the results of an expert evaluation study assessing the quality of the artificial clinical notes generated by the LLM. \pr{Next, Section \ref{sec:utility_spans} uses the extracted symptom spans to further analyze the content of the notes.} Following this, Section \ref{sec:CIE_baselines} evaluates the performance of simple symptom prediction models applied to both the tabular and textual portions of the dataset. Finally, Section \ref{sec:intended_use} explores the broader research applications of the dataset, including but not limited to clinical information extraction.

\subsection{Expert evaluation \pr{of generated notes}} \label{sec:expert_evaluation}

We asked five general practitioners to evaluate the quality of a random sample of $30$ generated notes. \rebut{The primary aim of our evaluation study was to assess whether the generated notes adhered to the instructions provided in the prompt. As a secondary aim, we sought to gain an indication of how realistic the content of the notes appeared.} \rebut{We invited general practitioners with an affiliation to the Department of Primary Care at our institution to take part in the evaluation study. The participants were recruited via email, and included based on their willingness to participate when provided information about the task's duration}. 
The evaluators got to see the two versions of the generated note (normal and compact), as well as the prompt that was used to generate them. 

\subsubsection{\pr{Evaluation criteria}}

\pr{The evaluators were asked to rate four criteria for each clinical note:}
\begin{enumerate}
    \item \textbf{Consistency}: The description of the patient's symptoms and underlying health conditions must correspond with the instructions provided in the prompt.
    \item \textbf{Realism}: The context and details invented by the LLM and added to the ``history'' portion of the note should be realistic given the symptoms experienced by the patient, as detailed in the prompt. Furthermore, the elements that are checked in the ``physical examination'' must be realistic in light of the patient information described in the ``history''. Both aspects are evaluated separately, focusing on content (what information is included in the note), rather than format (how the information is written down). 
    \item \textbf{Clinical accuracy}: The findings described in the ``physical examination'' must be clinically accurate, both in a standalone fashion and in relation to the patient's symptoms described in the ``history''. 
    \item \textbf{Quality of compact version}: The content of the compact version of the note must correspond well with the original note, and remain readable despite the use of abbreviations and shortcuts. We evaluate both aspects separately.
\end{enumerate}

To measure consistency, the evaluators were asked to assign a penalty for every element in the note that does not correspond with the requested information in the prompt, and for every symptom requested in the prompt that was missing from the generated note. These penalties were then turned into ratings from 1 (\textgreater 3 penalties) to 5 (0 penalties). 
All other aspects were directly rated on a scale of 1 (very bad) to 5 (perfect). \rebut{To ensure the clarity and intuitiveness of the evaluation criteria, we conducted a pilot run of the evaluation process and obtained feedback from a general practitioner who did not participate in the final study.}
For more details on the meaning of each rating in each criterion, as well as inter-annotator agreement scores, we refer to Appendix \ref{sec:app_evaluation}.\\

\subsubsection{\pr{Evaluation results}} 

\begin{table}[t]
\centering
\caption{Results of the expert evaluation study. We report average scores (ranging from 1-5) over $30$ notes, together with their standard deviation over the five evaluators.}
\begin{tabular}{ccccccc}
    \toprule
    & \multicolumn{4}{c}{\textbf{Normal}} & \multicolumn{2}{c}{\textbf{Compact}} \\
    \cmidrule(r){2-5} 
    \cmidrule(r){6-7}
    & \makecell[ct]{Consis-\\tency} & \makecell[ct]{Realism \\ (hist)} & \makecell[ct]{Realism \\ (phys)} & \makecell[ct]{Clinical \\ accuracy} & Content & \makecell[ct]{Reada-\\bility} \\
    \cmidrule(r){2-5} 
    \cmidrule(r){6-7}
    mean & 4.69 & 4.53 & 4.15 & 4.92 & 4.88 & 4.02 \\
    std & 0.12 & 0.21 & 0.30 & 0.07 & 0.10 & 0.31 \\
    \bottomrule
    \bottomrule
\end{tabular}
\label{tab:evaluation_results}
\end{table}

Table \ref{tab:evaluation_results} shows the results of the evaluation study. For each rated criterion, we calculate the average score over all $30$ notes, and report the mean and standard deviation over the five evaluators.
Our artificial notes were rated as highly consistent with the prompt, and therefore with the information present in the tabular portion of the dataset. As shown in Appendix \ref{sec:app_evaluation_results}, a large majority of the inconsistencies arise due to a violation of the additional instructions (usually by inventing additional symptoms), while the key information included in the prompt was still conveyed correctly in the note. High consistency between the tabular variables and the concepts mentioned in the text makes our dataset a reliable resource for information extraction tasks. \\

The evaluators also deem the notes sufficiently realistic, though the realism of the history section is rated higher on average than the realism of the physical examination. Out of those notes that scored worse, many included a clinical test that seemed unnecessary to the evaluators, while a few forgot a test that was deemed important. At the same time, the very high score for clinical accuracy is an important indication that the notes do not contain falsehoods. Multiple evaluators mentioned that while the content of the notes seemed realistic, the format did not, as their own notes would be more complex as opposed to the artificial notes, which use clean language and full sentences. This underscores the fact that the dataset should not be used to train any systems which will later be deployed on real notes, and should instead fulfill the role of a research benchmark only. \\

The compact versions of the notes score very well on content, mostly conveying the same information as the original. They score a little lower in terms of readability, which evaluators often attributed to the extensive use of abbreviations.\\

\subsection{\pr{Analysis of generated notes with symptom spans}} \label{sec:utility_spans}

\begin{table*}[t]
\caption{Descriptive statistics on the symptoms mentioned in the notes, calculated via the extracted symptom spans.}
\begin{subtable}[c]{\textwidth}
\centering
\caption{Statistics on the average number of spans per note, distribution of spans over symptoms, average length of spans, and location of spans (``history'' vs. ``physical examination'' portion of the note). For each metric, we report normal / compact.}
\resizebox{\textwidth}{!}{
\begin{tabular}{ccccccc}
    \toprule
    & \makecell[ct]{\textbf{Dyspnea}} & \makecell[ct]{\textbf{Cough}} & \makecell[ct]{\textbf{Pain}} & \makecell[ct]{\textbf{Nasal}} & \textbf{Fever} & \textbf{All} \\
    \cmidrule(r){2-6} 
    \cmidrule(r){7-7}
    \textbf{Avg. \# spans / note} & 0.96 / 0.90 & 1.03 / 0.96 & 0.52 / 0.49 & 0.85 / 0.81 & 0.79 / 0.74 & 4.16 / 3.90 \\
    \textbf{Total \% of spans} & 23.2\% / 23.1\% & 24.9\% / 24.6\% & 12.6\% / 12.5\% & 20.4\% / 20.7\% & 19.0\% / 19.0\% & 100\% / 100\% \\
    \textbf{Avg. \# words / span} & 8.34 / 4.86 & 8.83 / 5.22 & 9.34 / 5.96 & 8.92 / 5.98 & 7.04 / 4.07 & 8.46 / 5.17 \\
    \textbf{Avg. \# chars / span} & 57.21 / 30.45 & 56.70 / 32.14 & 63.54 / 36.57 & 63.05 / 41.74 & 44.82 / 23.95 & 56.71 / 32.73 \\
    \textbf{\% of spans in \textit{hist}} & 79.3\% / 78.8\% & 92.5\% / 91.4\% & 81.6\% / 80.8\% & 63.0\% / 62.1\% & 65.9\% / 65.2\% & 76.8\% / 76.1\% \\
    \textbf{\% of spans in \textit{phys}} & 20.7\% / 21.2\% & 8.5\% / 8.6\% & 18.4\% / 19.2\% & 37.0\% / 37.9\% & 34.1\% / 34.8\% & 23.2\% / 23.9\% \\
    \bottomrule
\end{tabular}
}
\vspace{2mm}
\label{tab:span_statistics_general}
\end{subtable}
\begin{subtable}[c]{\textwidth}
\centering
\caption{We show the percentage of notes where a symptom is mentioned (meaning there is at least one span mentioning the symptom in the note), and the average number of spans per note in which a symptom is mentioned. These are reported conditional on the value of the symptom (whether it is present in the patient or not).}
\resizebox{\textwidth}{!}{
\begin{tabular}{ccccccccccccc}
    \toprule
    & & \multicolumn{2}{c}{\textbf{Dyspnea}} & \multicolumn{2}{c}{\textbf{Cough}} & \multicolumn{2}{c}{\textbf{Pain}} & \multicolumn{2}{c}{\textbf{Nasal}} & \multicolumn{3}{c}{\textbf{Fever}}\\
    \cmidrule(r){3-4} \cmidrule(r){5-6} \cmidrule(r){7-8} \cmidrule(r){9-10} \cmidrule(r){11-13}
    & & yes & no & yes & no & yes & no & yes & no & high & low & none \\
    \cmidrule(r){3-4} \cmidrule(r){5-6} \cmidrule(r){7-8} \cmidrule(r){9-10} \cmidrule(r){11-13}
    \makecell[ct]{\textbf{\% of notes where} \\ \textbf{sympt. mentioned} \\ \textbf{($\ge$ 1 span / note)}\vspace{2mm}} & \makecell[ct]{Normal \\ Compact} & \makecell[ct]{99.1\% \\ 98.1\%} & \makecell[ct]{66.5\% \\ 62.5\%} & \makecell[ct]{97.0\% \\ 95.4\%} & \makecell[ct]{64.5\% \\ 59.4\%} & \makecell[ct]{81.7\% \\ 79.4 \%} & \makecell[ct]{39.9\% \\ 37.1\%} & \makecell[ct]{95.7\% \\ 95.4\%} & \makecell[ct]{26.6\% \\ 25.1\%} & \makecell[ct]{95.9\% \\ 95.7\%} & \makecell[ct]{71.6\% \\ 71.4\%} & \makecell[ct]{44.4\% \\ 42.4\%} \\
    \makecell[ct]{\textbf{Avg. \# spans where} \\ \textbf{sympt. mentioned / note}} & \makecell[ct]{Normal \\ Compact} & \makecell[ct]{2.03 \\ 1.91} & \makecell[ct]{0.70 \\ 0.65} & \makecell[ct]{1.78 \\ 1.66} & \makecell[ct]{0.65 \\ 0.60} & \makecell[ct]{1.28 \\ 1.21} & \makecell[ct]{0.40 \\ 0.37} & \makecell[ct]{2.47 \\ 2.38} & \makecell[ct]{0.30 \\ 0.28} & \makecell[ct]{2.46 \\ 2.33} & \makecell[ct]{1.55 \\ 1.48} & \makecell[ct]{0.52 \\ 0.48} \\
    \bottomrule
\end{tabular}
}
\label{tab:span_statistics_mentioned}
\end{subtable}
\label{tab:span_statistics}
\end{table*}

\pr{The symptom spans extracted in Section \ref{sec:span_extraction} allow us to quantify the presence of each of the 5 symptoms in the notes. We first evaluate the accuracy of the extracted spans with a small manual evaluation, after which we use the spans to analyze the content of the notes. \\}

\textbf{Manual evaluation}\quad To assess the quality of the automated span extraction process, we conducted an evaluation study over 100 randomly selected notes. We manually annotated each of these notes with phrases that were missed and phrases that were extracted incorrectly. From this, we calculated precision (how many extracted phrases are correct, over the total number of extracted phrases per note) and recall (how many extracted phrases are correct, over the total number of phrases that should have been extracted per note). Averaging these metrics over this set of 100 notes yields a precision of $94.06\%$ and recall of $98.78\%$ for the normal notes, and a precision of $93.55\%$ and recall of $94.50\%$ for the compact notes. \\

\textbf{Descriptive statistics}\quad The extracted spans allow us to calculate statistics on the presence of each of the 5 symptoms in the notes, which are reported in Table \ref{tab:span_statistics}. We highlight some of the most interesting findings. Table \ref{tab:span_statistics_general} shows that on average, less spans were extracted for the compact notes than for the normal notes. The explanation is probably two-fold: while some descriptions could be missing from the compact notes entirely, their challenging nature also means more mistakes will be made in the automated extraction, leading to some spans being missed (as reflected by the lower average recall in our manual evaluation). The symptom \textit{pain} is mentioned much less than the other 4 symptoms, even when it is present in the patient. Unsurprisingly, the average span length is much shorter in the compact notes, both in terms of words and characters. According to the extracted spans, most of the symptoms are described in the ``history'' portion of the clinical note, with proportions differing considerably between symptoms. The percentages of symptoms being mentioned at least once in the note (see Table \ref{tab:span_statistics_mentioned}) correspond quite well with the probabilities used to build the prompt, as defined in Section \ref{sec:prompting_strategy}. The average number of times a symptom is mentioned per note is much lower when the symptom is not present in the patient. In total, 16.3\% of all normal notes do not mention any of the 5 symptoms at all (no spans extracted), while this percentage is 16.4\% for the compact notes. 

\subsection{Symptom predictor baselines} \label{sec:CIE_baselines}


In order to set a baseline for future information extraction tasks, such as the one envisioned in Fig. \ref{fig:IE_with_background}, we run various prediction models on both the tabular and textual parts of the dataset. These models are trained to predict each of the five symptoms: \textit{dyspnea}, \textit{cough}, \textit{pain}, \textit{fever} and \textit{nasal}. \pr{Symptom prediction is a useful task to show the utility of our dataset, since information about the symptoms can be inferred from both the tabular portion of the dataset (through the other variables), as well as the textual portion of our dataset (the clinical notes which partially describe the symptoms).} \\

\pr{Section \ref{sec:task_overview} describes the task setup and gives an overview of the various types of baselines we run on our dataset.} Section \ref{sec:models} provides a detailed overview of each baseline model, followed by Section \ref{sec:results}, which analyzes their performance on the task of automated symptom extraction using the SimSUM dataset.

\subsubsection{\pr{Task overview}} \label{sec:task_overview}
\noindent
\pr{\textbf{Tabular-only baselines}} \quad Two of our baselines only get to see the tabular portion of the dataset at the input: Bayesian network (\textbf{BN-tab}) and XGBoost (\textbf{XGBoost-tab}). \rebut{We include XGBoost because it has consistently demonstrated state-of-the-art performance on a wide range of tabular prediction tasks, often serving as a strong baseline in machine learning benchmarks \citep{xgboost_tabular}.} We use both models to predict each symptom in three settings, differing from one another in the set of tabular features that are taken as an input, which we call the evidence:
\setlist{nolistsep}
\begin{itemize}[noitemsep]
    \item $\mathcal{P}(\text{\textit{sympt}} \mid \text{\textit{all}})$: Predict the symptom given all other tabular features as evidence. This set includes the background, diagnoses, non-clinical, treatment and outcome variables, as well as the other symptoms.
    \item $\mathcal{P}(\text{\textit{sympt}} \mid \text{\textit{no-sympt}})$: Predict the symptom given all other tabular features as evidence, except for the other symptoms. This mimics the setting where we have tabular features available in the patient record, but have not extracted any symptoms from the text yet. This set includes the background, diagnoses, non-clinical, treatment and outcome variables. 
    \item $\mathcal{P}(\text{\textit{sympt}} \mid \text{\textit{realistic}})$: Predict the symptom given a more realistic set of tabular features as evidence. We do not expect \textit{policy}, \textit{self{\text-}employed} and \textit{\pr{days at home}} to be recorded in any kind of realistic patient record, and therefore leave them out of this evidence set. As in the \textit{no-sympt} setting, we do not include the symptoms either. In other words, this set includes the background, diagnoses, season and treatment variables. \\
\end{itemize}

\noindent
\textbf{\pr{Text baselines}} \quad Apart from the tabular-only baselines, we also train baselines that get to see the text. Our \textbf{neural-text} classifier takes only the text as an input (in the form of a pretrained clinical sentence embedding) and outputs the probability that a symptom is mentioned in the text. We extend this text-only baseline by concatenating a numerical representation of the tabular features to the text embedding at the input, forming the \textbf{neural-text-tab} baseline. Again, we do this for each of the three evidence settings outlined above. Note that this is the only model that combines both the background knowledge available in the tabular features with the unstructured text, and it does so in a naive way. Future work will focus on improving the performance of this model by exploiting the relations between the tabular concepts, as envisioned in Fig. \ref{fig:IE_with_background}. \\

\subsubsection{Models} \label{sec:models}

\noindent
\textbf{BN-tab}\quad We provide the causal structure in Fig. \ref{fig:data_generation} to the BN, and learn all parameters in the conditional probability tables (CPTs), Noisy-OR distributions, logistic regression model and Poisson regression model from the training data. In each case, we use maximum likelihood estimation to estimate the parameters. Where we \rebut{do not} directly learn a CPT (for the variables \textit{dyspnea}, \textit{cough}, \textit{pain}, \textit{nasal}, \textit{antibiotics} and \textit{\pr{days at home}}), we evaluate the learned distribution for each combination of child and parent values to obtain a CPT. For more details, we refer to Appendix \ref{app:BN-tab}. We then use variable elimination over the full joint distribution to evaluate the capability of the learned BN to predict each of the symptoms, taking different sets of variables as evidence according to the three settings described earlier (\textit{all}, \textit{no-sympt} and \textit{realistic}). \\

\noindent
\textbf{XGBoost-tab}\quad We train an XGBoost classifier for each symptom in combination with each of the three evidence settings, meaning each classifier sees a different set of tabular features at the input. We optimize the hyperparameters separately for each combination (15 in total) using 5-fold cross-validation. For more details, we refer to Appendix \ref{app:XGBoost-tab}. \\

\noindent
\textbf{Neural-text}\quad We train a neural classifier that takes only the text as an input and is trained to predict the probability a symptom is mentioned. We train separate classifiers for each symptom. We first split the text into sentences, and transform these into an embedding using the pretrained clinical representation model BioLORD-2023 \citep{biolord}. We explore four settings for turning these sentence embeddings into a single note embedding: 
\begin{itemize}
    \item \textit{hist}: We average all sentence embeddings for the sentences in the ``history'' portion of the note. 
    \item \textit{phys}: We average all sentence embeddings for the sentences in the ``physical examination'' portion of the note.
    \item \textit{mean}: To get a single representation for the full note, we take the average of the \textit{hist} and \textit{phys} embeddings. 
    \item \textit{concat}: Idem as previous, but now the embeddings for the two portions are concatenated.
\end{itemize}

\noindent
Additionally, we create symptom-specific \textit{span} embeddings for each note as well. To this end, we take the set of spans mentioning the symptom in each note (as extracted automatically in Section \ref{sec:span_extraction}), transform each of these spans into their BioLORD embedding, and take the average to get one span embedding per symptom per note. 
No matter the type, each note embedding is fed into a multi-layer perceptron with one hidden layer, followed by a Sigmoid activation for the symptoms \textit{dyspnea}, \textit{cough}, \textit{pain} and \textit{nasal}, and a Softmax activation with 3 outputs heads for the symptom $fever$. We trained each model (i.e. each combination of symptom and embedding type) using the binary or multiclass cross-entropy objective over the symptom labels. For more details, we refer to Appendix \ref{app:neural-text}. \\

\noindent
\textbf{Neural-text-tab} \quad We extend the \textbf{neural-text} baseline by concatenating the \textit{mean} text embeddings with the tabular variables at the input of each neural classifier. All categorical tabular variables were first transformed to a one-hot encoding, while the variable \textit{\pr{days at home}} was preprocessed using standard scaling. We used the same architecture as the \textbf{neural-text} baseline (only changing the dimension of the input layer), and again trained separate classifiers for each symptom combined with each evidence setting (\textit{all}, \textit{no-sympt} and \textit{realistic}). For more details, we refer to Appendix \ref{app:neural-text-tab}.\\

\noindent
\textbf{\pr{Training} setup} \quad We use a random $8{,}000$/$2{,}000$ split to obtain a train and test subset of SimSUM. We use cross-validation on the train set to tune any hyperparameters associated with each model, and report the average F1-score over the test set after training with 5 different seeds. \rebut{Seeding randomizes the initialization of the model weights, as well as batching.} For the binary symptoms, we use a $0.5$ decision threshold. For fever, which has three possible categories, the class with the highest predicted probability is chosen. 

\subsubsection{\rebut{Symptom predictor results}} \label{sec:results}

\begin{table}[t]
\centering
\caption{F1-score obtained over the test set for each of our baseline models. For fever, we report the macro F1 score over all three categories. The results for the text classifiers trained over the normal vs. the compact version of the notes are grouped together for readability. We report results for the \textit{mean} embedding type, while \textbf{neural-text} baseline results for the other embedding types can be found in Table \ref{tab:text_classifier_results}. We \underline{underline} the best result obtained by the tabular-only models, while the best overall result per symptom and note version is in shown in \textbf{bold}.}
\begin{tabular}{lccccc}
    \toprule
    & \textbf{dyspnea} & \textbf{cough} & \textbf{pain} & \textbf{nasal} & \textbf{fever} \\
    \midrule
    \textbf{BN-tab} &&&&& \\
    - all & \underline{0.7370} & 0.7816 & 0.2386 & \underline{0.7146} & \underline{0.4864} \\
    - no-sympt & 0.7153 & 0.7776 & 0.1312 & \underline{0.7146} & 0.4384 \\
    - realistic & 0.6698 & 0.7763 & 0.0280 & \underline{0.7146} & 0.3594 \\
    \textbf{XGBoost-tab} &&&&&\\
    - all & 0.6639 & \underline{0.7848} & \underline{0.4070} & 0.7130 & 0.4111 \\
    - no-sympt & 0.6612	& 0.7779 & 0.3638 & \underline{0.7146} & 0.4015 \\
    - realistic & 0.6626 & 0.7798 & 0.3698 & \underline{0.7146}	& 0.3951 \\
    \midrule
    \textbf{neural-text} & & & & & \\
    - normal & \textbf{0.9617} & \textbf{0.9603} & 0.8143 & \textbf{0.9628} & \textbf{0.9096} \\
    \textbf{neural-text-tab} & & & & & \\
    - normal + all & 0.9544	& 0.9555 & 0.8191 & 0.9613 & 0.9091 \\
    - normal + no-sympt & 0.9594 & 0.9577 & 0.8155 & 0.9611	& 0.8996 \\
    - normal + realistic & 0.9569 & 0.9596 & \textbf{0.8266}	& 0.9591 & 0.9029 \\
    \midrule
    \textbf{neural-text} & & & & & \\
    - compact & 0.9444 & 0.9397 & \textbf{0.7940} & 0.9622 & 0.9010 \\
    \textbf{neural-text-tab} & & & & & \\
    - compact + all & \textbf{0.9532} & 0.9395 & 0.7873 & 0.9589 & 0.8962 \\
    - compact + no-sympt & 0.9429 & 0.9456 & 0.7856 & \textbf{0.9632} & 0.8973 \\
    - compact + realistic & 0.9379 & \textbf{0.9480} & 0.7922 & 0.9630 & \textbf{0.9025} \\
    \bottomrule
    \bottomrule
\end{tabular}
\label{tab:baseline_results}
\end{table}

\begin{table}[t]
\centering
\caption{Results for the \textbf{neural-text} baseline using different embedding types at the input. We report F1-score over the test set for both the normal and compact versions of the notes. The best results per symptoms and per note version is shown in \textbf{bold}, the runner-up is underlined. Fig. \ref{fig:text_emb_results} in the Appendix compares these results visually.
}
\begin{tabular}{lccccc}
    \toprule
    \textbf{Normal}& \textbf{dyspnea} & \textbf{cough} & \textbf{pain} & \textbf{nasal} & \textbf{fever} \\
    \midrule
    Sentence embedding &&&&& \\
    - hist & 0.9520 & \underline{0.9652} & 0.8089 & 0.9504 & 0.9040 \\
    - phys & 0.8969	& 0.8226 & 0.6956 & 0.9565 & 0.8523 \\
    - mean & \underline{0.9617} & 0.9603 & 0.8143 & \underline{0.9628} & \underline{0.9096} \\
    - concat & 0.9504 & 0.9572 & \underline{0.8268} & 0.9589 & 0.9044 \\
    Span embedding & \textbf{0.9730} & \textbf{0.9785} & \textbf{0.8731} & \textbf{0.9712} & \textbf{0.9226} \\
    \midrule
    \midrule
    \textbf{Compact}& \textbf{dyspnea} & \textbf{cough} & \textbf{pain} & \textbf{nasal} & \textbf{fever} \\
    \midrule
    Sentence embedding &&&&& \\
    - hist & 0.9307	& \underline{0.9523} & 0.7734 & 0.9515 & 0.9001 \\
    - phys & 0.8824	& 0.7802 & 0.6514 & 0.9531 & 0.8315 \\
    - mean & 0.9444 & 0.9397 & \underline{0.7940} & \underline{0.9622} & \underline{0.9010} \\
    - concat & \underline{0.9522} & 0.9473 & 0.7934 & 0.9411 & 0.9007 \\
    Span embedding & \textbf{0.9685} & \textbf{0.9719} & \textbf{0.8613} & \textbf{0.9705} & \textbf{0.9207} \\
    \bottomrule
\end{tabular}
\label{tab:text_classifier_results}
\end{table}

Table \ref{tab:baseline_results} compares the results obtained for all baselines, using the \textit{mean} embedding type for the text-based classifiers. The tabular-only baselines (\textbf{BN-tab} and \textbf{XGBoost-tab}) perform consistently worse than the baselines that include text (\textbf{neural-text} and \textbf{neural-text-tab}). The evidence setting where \textit{all} other features are included as evidence usually performs better than the other settings. 

For the text baselines, we note that there is little room for improvement in the dyspnea, cough and nasal classifiers. On the other hand, the symptoms pain and fever are harder to predict. 
We can assume that this is largely because these symptoms are mentioned less often in the notes, even if they are present in the patient, as supported by Table \ref{tab:span_statistics}.
We also see a consistent gap in performance between the normal and compact notes, which can be attributed to the higher complexity of the latter. While the \textbf{neural-text-tab} baseline does not manage to outperform the \textbf{neural-text} baseline for the normal notes, we do see marginal improvements by including the tabular features in the case of the compact notes. 

Table \ref{tab:text_classifier_results} further breaks down the results for the \textbf{neural-text} classifier over the different embedding types. We clearly see that the \textit{span} embeddings perform best, since this method preselects phrases that are important for the prediction of a certain symptom. While the same phrases are still implicitly embedded in the other types, the important information gets diluted as sentence embeddings are averaged, and might even be lost entirely. However, in a realistic setting, one would rarely have access to these symptom-specific \textit{span} embeddings, as this would require either a manual annotation effort, or a (small) local LLM to facilitate automated annotation in a way that preserves patient privacy. We do not expect such a local LLM to achieve near the same span detection performance as GPT-4o, which we were only able to use in this case due to the artificial nature of our notes. The performance achieved with these \textit{span} embeddings should therefore be seen \rebut{as a best-case reference within our benchmark, rather than an achievable goal in a practical setting.}

Comparing the four different sentence embedding types, the performance difference between the \textit{mean} and \textit{concat} settings is usually small, with \textit{mean} slightly outperforming \textit{concat} overall. While the score for \textit{hist} comes close to those for \textit{mean} and \textit{concat}, the latter two usually still outperform the former for the normal notes, showing that there is some complementary information in the ``history'' and ``physical examination'' portions. At the same time, we note that using only the \textit{hist} embedding clearly outperforms the \textit{phys}-only setting. This makes sense, as the ``history'' section of the note outlines the symptoms experienced by the patient more clearly. 

\subsection{Intended uses} \label{sec:intended_use}

\rebut{The primary purpose of the SimSUM dataset is to enable research on clinical information extraction in multimodal contexts, where effective integration of structured and unstructured data is required.}
A key objective for future work is to realize the approach illustrated in Fig. \ref{fig:IE_with_background}, in which domain knowledge links structured features to textual concepts, enabling more accurate information extraction. As demonstrated in the previous section, even a simple method—concatenating tabular features with text embeddings at the input of a neural classifier—already improves symptom extraction performance, particularly on the more challenging, compact notes. This suggests that background information can help recover symptoms that are mentioned less explicitly in the text, underscoring the utility of our dataset for developing and evaluating such approaches.

Apart from this, we also foresee multiple secondary uses of the dataset. First, the dataset could facilitate research on the automation of clinical reasoning over tabular data and text, following \rebut{our previous work in} \citet{rabaey2024clinical}. Second, it could be used to benchmark causal effect estimation methods in the presence of textual confounders, as proposed by \citet{text_CATE} \rebut{and \citet{time_text_confounding}}. This is possible thanks to the purposeful inclusion of both a treatment and outcome variable in our dataset. Third, there has been increasing interest in clinical synthetic data \citep{hernandez2022synthetic}, where a set of patient characteristics is turned into a synthetic version that is meant to protect the privacy of individuals in the original dataset. Our dataset could serve as a benchmark for comparing synthetic data generation methods that jointly generate tabular variables and text \rebut{\citep{EHR_generation, synthetic_EMR_text, promptehr, generative_AI_synthetic}}. In short, any area of research focusing on the intersection of tabular data and text in healthcare can potentially benefit from our proposed benchmark.

Conversely, there are important limitations to the intended use of SimSUM. As it is fully synthetic—constructed from expert-defined tabular data and language model–generated notes—it does not capture the full complexity, variability, or noise characteristic of real electronic health records. For this reason, we strongly advise against using SimSUM to evaluate clinical model performance or to train models intended for production. Instead, its value lies in supporting research scenarios, such as those outlined above, which benefit from the availability of a known and controllable ground-truth structure.

\section{Conclusions} \label{sec:discussion}


We introduced SimSUM, a simulated dataset of structured and unstructured medical records designed to support research on clinical information extraction using background tabular data. By explicitly linking structured features and textual concepts through a Bayesian network, SimSUM provides a controlled setting to explore how domain knowledge can enhance information extraction from clinical notes. Indeed, our baseline results suggest that integrating tabular background features with text improves symptom extraction, particularly for harder-to-predict cases like pain and fever. Future research can explore more advanced hybrid models that leverage domain knowledge to link structured and unstructured data. Additionally, the annotated symptom spans, though imperfect, offer a foundation for developing lightweight span detection models and explainable extraction methods.\\



\noindent
\textbf{Limitations} \quad Despite aiming for realism, SimSUM intentionally simplifies many aspects of real-world clinical data to support its role as a controlled research benchmark. The dataset is synthetic by design, with known ground-truth relationships, and is not intended to generalize to real clinical notes or settings. Expert reviewers confirmed that while the content of the notes is generally plausible, the writing style does not reflect how clinicians typically document patient encounters. Additionally, SimSUM represents static patient snapshots rather than evolving time series, which limits its applicability in domains like intensive care where temporal dynamics are critical. 

\backmatter

\section*{Declarations}

\paragraph{Ethics approval and consent to participate} Not applicable.

\paragraph{Consent for publication} Not applicable. 

\paragraph{Availability of data and materials} The SimSUM dataset proposed during the current study is freely available on Github, at \url{https://github.com/prabaey/SimSUM}. This repository also contains all code necessary to reproduce the results presented in this work. 

\paragraph{Competing interests} The authors declare that they have no competing interests. 

\paragraph{Funding} Paloma Rabaey's research is funded by the Research Foundation Flanders (FWO Vlaanderen) with grant number 1170124N. This research also received funding from the Flemish government under the “Onderzoeksprogramma Artificiële Intelligentie (AI) Vlaanderen” programme.


\paragraph{Acknowledgements} The authors would like to thank Géraldine Deberdt, Thibault Detremerie, An De Sutter, Veerle Piessens and Florian Stul for participating in the expert evaluation of the artificial clinical notes. The authors would also like to thank Henri Arno for his early contributions to the dataset design.

\section*{\pr{List of abbreviations}}

\pr{The following is a list of abbreviations that were used throughout the work.}
\pr{\begin{itemize}
    \item \textbf{EHR}: Electronic health record
    \item \textbf{CIE}: Clinical information extraction
    \item \textbf{BN}: Bayesian network
    \item \textbf{DAG}: Directed acyclic graph
    \item \textbf{CPT}: Conditional probability table
    \item \textbf{LLM}: Large language model
\end{itemize}}

\bibliography{references}

@book{koller2009probabilistic,
  author = {Koller, D. and Friedman, N.},
  isbn = {9780262013192},
  publisher = {MIT Press},
  series = {Adaptive computation and machine learning},
  title = {Probabilistic Graphical Models: Principles and Techniques},
  year = 2009
}

@article{mimic3,
  title={MIMIC-III, a freely accessible critical care database},
  author={Johnson, Alistair EW and Pollard, Tom J and Shen, Lu and Lehman, Li-wei H and Feng, Mengling and Ghassemi, Mohammad and Moody, Benjamin and Szolovits, Peter and Anthony Celi, Leo and Mark, Roger G},
  journal={Scientific data},
  volume={3},
  number={1},
  pages={1--9},
  year={2016},
  publisher={Nature Publishing Group}
}

@article{mimic4,
  title={MIMIC-IV, a freely accessible electronic health record dataset},
  author={Johnson, Alistair EW and Bulgarelli, Lucas and Shen, Lu and Gayles, Alvin and Shammout, Ayad and Horng, Steven and Pollard, Tom J and Hao, Sicheng and Moody, Benjamin and Gow, Brian and others},
  journal={Scientific data},
  volume={10},
  number={1},
  pages={1},
  year={2023},
  publisher={Nature Publishing Group UK London}
}

@inproceedings{biodex,
    title = "{B}io{DEX}: Large-Scale Biomedical Adverse Drug Event Extraction for Real-World Pharmacovigilance",
    author = "D{'}Oosterlinck, Karel  and
      Remy, Fran{\c{c}}ois  and
      Deleu, Johannes  and
      Demeester, Thomas  and
      Develder, Chris  and
      Zaporojets, Klim  and
      Ghodsi, Aneiss  and
      Ellershaw, Simon  and
      Collins, Jack  and
      Potts, Christopher",
    booktitle = "Findings of the Association for Computational Linguistics: EMNLP 2023",
    month = dec,
    year = "2023",
    pages = "13425--13454",
}

@article{tcga_reports,
  title={TCGA-Reports: A machine-readable pathology report resource for benchmarking text-based AI models},
  author={Kefeli, Jenna and Tatonetti, Nicholas},
  journal={Patterns},
  volume={5},
  number={3},
  year={2024},
  publisher={Elsevier}
}

@InProceedings{pgmpy,
  author    = { {A}nkur {A}nkan and {A}binash {P}anda },
  title     = { pgmpy: {P}robabilistic {G}raphical {M}odels using {P}ython },
  booktitle = { {P}roceedings of the 14th {P}ython in {S}cience {C}onference },
  pages     = { 6 - 11 },
  year      = { 2015 },
}

@misc{GPT4o,
author={OpenAI}, 
title={{Models -- GPT-4o}},
howpublished="\url{https://platform.openai.com/docs/models/gpt-4o}",
year={2024},
note="{Online; accessed 12 August 2024}"
}

@article{extracting_information_EMR,
    author = {Ford, Elizabeth and Carroll, John A and Smith, Helen E and Scott, Donia and Cassell, Jackie A},
    title = "{Extracting information from the text of electronic medical records to improve case detection: a systematic review}",
    journal = {J Am Med Inform Assoc},
    volume = {23},
    number = {5},
    pages = {1007-1015},
    year = {2016},
    issn = {1067-5027},
}

@article{CDS_infectious_diseases,
title = {Machine learning for clinical decision support in infectious diseases: {A} narrative review of current applications},
journal = {Clin Microbiol Infect},
volume = {26},
number = {5},
pages = {584-595},
year = {2020},
issn = {1198-743X},
author = {N. Peiffer-Smadja and T.M. Rawson and R. Ahmad and A. Buchard and et al.}
}

@article{clinical_text_classification,
title = {Clinical text classification research trends: Systematic literature review and open issues},
journal = {Expert Syst Appl},
volume = {116},
pages = {494-520},
year = {2019},
issn = {0957-4174},
author = {Ghulam Mujtaba and Liyana Shuib and Norisma Idris and Wai Lam Hoo and Ram Gopal Raj and Kamran Khowaja and Khairunisa Shaikh and Henry Friday Nweke}
}

@article{combining_structured_unstructured,
   Author="Zhang, D.  and Yin, C.  and Zeng, J.  and Yuan, X.  and Zhang, P. ",
   Title="{{C}ombining structured and unstructured data for predictive models: {A} deep learning approach}",
   Journal="BMC Med Inform Decis Mak",
   Year="2020",
   Volume="20",
   Number="1",
   Pages="280",
}

@inproceedings{multimodal_ICD,
  title={Multimodal machine learning for automated ICD coding},
  author={Xu, Keyang and Lam, Mike and Pang, Jingzhi and Gao, Xin and Band, Charlotte and Mathur, Piyush and Papay, Frank and Khanna, Ashish K and Cywinski, Jacek B and Maheshwari, Kamal and others},
  booktitle={Machine learning for healthcare conference},
  pages={197--215},
  year={2019},
  organization={PMLR}
}

@article{medBERT,
  title={Med-BERT: pretrained contextualized embeddings on large-scale structured electronic health records for disease prediction},
  author={Rasmy, Laila and Xiang, Yang and Xie, Ziqian and Tao, Cui and Zhi, Degui},
  journal={NPJ digital medicine},
  volume={4},
  number={1},
  pages={86},
  year={2021},
  publisher={Nature Publishing Group UK London}
}

@article{BEHRT,
  title={BEHRT: transformer for electronic health records},
  author={Li, Yikuan and Rao, Shishir and Solares, Jos{\'e} Roberto Ayala and Hassaine, Abdelaali and Ramakrishnan, Rema and Canoy, Dexter and Zhu, Yajie and Rahimi, Kazem and Salimi-Khorshidi, Gholamreza},
  journal={Scientific reports},
  volume={10},
  number={1},
  pages={7155},
  year={2020},
  publisher={Nature Publishing Group UK London}
}

@article{multimodal_matters,
  title={Multimodal data matters: language model pre-training over structured and unstructured electronic health records},
  author={Liu, Sicen and Wang, Xiaolong and Hou, Yongshuai and Li, Ge and Wang, Hui and Xu, Hui and Xiang, Yang and Tang, Buzhou},
  journal={IEEE Journal of Biomedical and Health Informatics},
  volume={27},
  number={1},
  pages={504--514},
  year={2022},
  publisher={IEEE}
}

@article{clinicalBERT,
  title={Clinicalbert: Modeling clinical notes and predicting hospital readmission},
  author={Huang, Kexin and Altosaar, Jaan and Ranganath, Rajesh},
  journal={arXiv preprint arXiv:1904.05342},
  year={2019}
}

@article{clinicalT5,
  title={Clinical-t5: Large language models built using mimic clinical text},
  author={Lehman, Eric and Johnson, Alistair},
  journal={PhysioNet},
  year={2023}
}

@article{MedPalm,
  title={Large language models encode clinical knowledge},
  author={Singhal, Karan and Azizi, Shekoofeh and Tu, Tao and Mahdavi, S Sara and Wei, Jason and Chung, Hyung Won and Scales, Nathan and Tanwani, Ajay and Cole-Lewis, Heather and Pfohl, Stephen and others},
  journal={Nature},
  volume={620},
  number={7972},
  pages={172--180},
  year={2023},
  publisher={Nature Publishing Group}
}

@inproceedings{BioMistral,
    title = "{B}io{M}istral: A Collection of Open-Source Pretrained Large Language Models for Medical Domains",
    author = "Labrak, Yanis  and
      Bazoge, Adrien  and
      Morin, Emmanuel  and
      Gourraud, Pierre-Antoine  and
      Rouvier, Mickael  and
      Dufour, Richard",
    booktitle = "Findings of the Association for Computational Linguistics: ACL 2024",
    month = aug,
    year = "2024",
    address = "Bangkok, Thailand",
    publisher = "Association for Computational Linguistics",
    pages = "5848--5864"
}

@article{clinical_information_extraction,
  title={Clinical information extraction applications: a literature review},
  author={Wang, Yanshan and Wang, Liwei and Rastegar-Mojarad, Majid and Moon, Sungrim and Shen, Feichen and Afzal, Naveed and Liu, Sijia and Zeng, Yuqun and Mehrabi, Saeed and Sohn, Sunghwan and others},
  journal={Journal of biomedical informatics},
  volume={77},
  pages={34--49},
  year={2018},
  publisher={Elsevier}
}

@article{medical_information_extraction,
  title={Medical information extraction in the age of deep learning},
  author={Hahn, Udo and Oleynik, Michel},
  journal={Yearbook of medical informatics},
  volume={29},
  number={01},
  pages={208--220},
  year={2020},
  publisher={Georg Thieme Verlag KG}
}

@inproceedings{EHRCon,
 author = {Kwon, Yeonsu and Kim, Jiho and Lee, Gyubok and Bae, Seongsu and Kyung, Daeun and Cha, Wonchul and Pollard, Tom and Johnson, Alistair and Choi, Edward},
 booktitle = {Advances in Neural Information Processing Systems},
 title = {EHRCon: Dataset for Checking Consistency between Unstructured Notes and Structured Tables in Electronic Health Records},
 volume = {37},
 year = {2024}
}

@article{explainability_LLMs,
  title={Explainability for large language models: A survey},
  author={Zhao, Haiyan and Chen, Hanjie and Yang, Fan and Liu, Ninghao and Deng, Huiqi and Cai, Hengyi and Wang, Shuaiqiang and Yin, Dawei and Du, Mengnan},
  journal={ACM Transactions on Intelligent Systems and Technology},
  volume={15},
  number={2},
  pages={1--38},
  year={2024},
  publisher={ACM New York, NY}
}

@article{opportunities_challenges_biomedicine,
  title={Opportunities and challenges for ChatGPT and large language models in biomedicine and health},
  author={Tian, Shubo and Jin, Qiao and Yeganova, Lana and Lai, Po-Ting and Zhu, Qingqing and Chen, Xiuying and Yang, Yifan and Chen, Qingyu and Kim, Won and Comeau, Donald C and others},
  journal={Briefings in Bioinformatics},
  volume={25},
  number={1},
  pages={bbad493},
  year={2024},
  publisher={Oxford University Press}
}

@article{CML_for_healthcare,
  title={Causal machine learning for healthcare and precision medicine},
  author={Sanchez, Pedro and Voisey, Jeremy P and Xia, Tian and Watson, Hannah I and O’Neil, Alison Q and Tsaftaris, Sotirios A},
  journal={Royal Society Open Science},
  volume={9},
  number={8},
  pages={220638},
  year={2022},
  publisher={The Royal Society}
}

@article{black_box,
  title={The three ghosts of medical AI: Can the black-box present deliver?},
  author={Quinn, Thomas P and Jacobs, Stephan and Senadeera, Manisha and Le, Vuong and Coghlan, Simon},
  journal={Artificial intelligence in medicine},
  volume={124},
  pages={102158},
  year={2022},
  publisher={Elsevier}
}

@article{rudin2019stop,
  title={Stop explaining black box machine learning models for high stakes decisions and use interpretable models instead},
  author={Rudin, Cynthia},
  journal={Nature machine intelligence},
  volume={1},
  number={5},
  pages={206--215},
  year={2019},
  publisher={Nature Publishing Group UK London}
}

@article{explainable_trees,
  title={From local explanations to global understanding with explainable AI for trees},
  author={Lundberg, Scott M and Erion, Gabriel and Chen, Hugh and DeGrave, Alex and Prutkin, Jordan M and Nair, Bala and Katz, Ronit and Himmelfarb, Jonathan and Bansal, Nisha and Lee, Su-In},
  journal={Nature machine intelligence},
  volume={2},
  number={1},
  pages={56--67},
  year={2020},
  publisher={Nature Publishing Group}
}

@inproceedings{rabaey2024clinical,
  title={Clinical Reasoning over Tabular Data and Text with Bayesian Networks},
  author={Rabaey, Paloma and Deleu, Johannes and Heytens, Stefan and Demeester, Thomas},
  booktitle={International Conference on Artificial Intelligence in Medicine},
  pages={229--250},
  year={2024},
  organization={Springer}
}

@article{EHR_generation,
  title={Natural language generation for electronic health records},
  author={Lee, Scott H},
  journal={NPJ digital medicine},
  volume={1},
  number={1},
  pages={63},
  year={2018},
  publisher={Nature Publishing Group UK London}
}

@article{biolord,
    author = {Remy, François and Demuynck, Kris and Demeester, Thomas},
    title = "{BioLORD-2023: semantic textual representations fusing large language models and clinical knowledge graph insights}",
    journal = {Journal of the American Medical Informatics Association},
    pages = {ocae029},
    year = {2024},
    month = {02}
}

@inproceedings{text_CATE,
  title={From Text to Treatment Effects: A Meta-Learning Approach to Handling Text-Based Confounding},
  author={Arno, Henri and Rabaey, Paloma and Demeester, Thomas},
  booktitle={Causal Representation Learning Workshop at NeurIPS 2024},
  year={2024}
}

@article{synthetic_EMR_text,
  title={A method for generating synthetic electronic medical record text},
  author={Guan, Jiaqi and Li, Runzhe and Yu, Sheng and Zhang, Xuegong},
  journal={IEEE/ACM transactions on computational biology and bioinformatics},
  volume={18},
  number={1},
  pages={173--182},
  year={2019},
  publisher={IEEE}
}

@article{hernandez2022synthetic,
  title={{Synthetic data generation for tabular health records: A systematic review}},
  author={Hernandez, Mikel and Epelde, Gorka and Alberdi, Ane and Cilla, Rodrigo and Rankin, Debbie},
  journal={Neurocomputing},
  volume={493},
  pages={28--45},
  year={2022},
  publisher={Elsevier}
}

@book{krippendorff,
  title={Content analysis: An introduction to its methodology},
  author={Krippendorff, Klaus},
  year={2018},
  publisher={Sage publications}
}

@article{time_text_confounding,
  title={LLM-Driven Treatment Effect Estimation Under Inference Time Text Confounding},
  author={Ma, Yuchen and Frauen, Dennis and Schweisthal, Jonas and Feuerriegel, Stefan},
  journal={arXiv preprint arXiv:2507.02843},
  year={2025}
}

@inproceedings{promptehr,
  title={PromptEHR: Conditional electronic healthcare records generation with prompt learning},
  author={Wang, Zifeng and Sun, Jimeng},
  booktitle={Proceedings of the Conference on Empirical Methods in Natural Language Processing. Conference on Empirical Methods in Natural Language Processing},
  volume={2022},
  pages={2873},
  year={2022}
}

@article{generative_AI_synthetic,
  title={Generative AI for synthetic data across multiple medical modalities: A systematic review of recent developments and challenges},
  author={Ibrahim, Mahmoud and Al Khalil, Yasmina and Amirrajab, Sina and Sun, Chang and Breeuwer, Marcel and Pluim, Josien and Elen, Bart and Ertaylan, G{\"o}khan and Dumontier, Michel},
  journal={Computers in biology and medicine},
  volume={189},
  pages={109834},
  year={2025},
  publisher={Elsevier}
}

@article{LLMs_information_extraction,
  title={Large language models for generative information extraction: A survey},
  author={Xu, Derong and Chen, Wei and Peng, Wenjun and Zhang, Chao and Xu, Tong and Zhao, Xiangyu and Wu, Xian and Zheng, Yefeng and Wang, Yang and Chen, Enhong},
  journal={Frontiers of Computer Science},
  volume={18},
  number={6},
  pages={186357},
  year={2024},
  publisher={Springer}
}

@article{pretrained_biomedical,
  title={Pre-trained language models in biomedical domain: A systematic survey},
  author={Wang, Benyou and Xie, Qianqian and Pei, Jiahuan and Chen, Zhihong and Tiwari, Prayag and Li, Zhao and Fu, Jie},
  journal={ACM Computing Surveys},
  volume={56},
  number={3},
  pages={1--52},
  year={2023},
  publisher={ACM New York, NY}
}

@article{medical_informed_ML,
  title={Medical-informed machine learning: integrating prior knowledge into medical decision systems},
  author={Sirocchi, Christel and Bogliolo, Alessandro and Montagna, Sara},
  journal={BMC Medical Informatics and Decision Making},
  volume={24},
  number={Suppl 4},
  pages={186},
  year={2024},
  publisher={Springer}
}

@article{BN_healthcare,
  title={A comprehensive scoping review of Bayesian networks in healthcare: Past, present and future},
  author={Kyrimi, Evangelia and McLachlan, Scott and Dube, Kudakwashe and Neves, Mariana R and Fahmi, Ali and Fenton, Norman},
  journal={Artificial Intelligence in Medicine},
  volume={117},
  pages={102108},
  year={2021},
  publisher={Elsevier}
}

@article{medical_KG,
  title={Medical knowledge graph: Data sources, construction, reasoning, and applications},
  author={Wu, Xuehong and Duan, Junwen and Pan, Yi and Li, Min},
  journal={Big data mining and analytics},
  volume={6},
  number={2},
  pages={201--217},
  year={2023},
  publisher={TUP}
}

@article{improving_ehr,
  title={Improving the quality and utility of electronic health record data through ontologies},
  author={Lin, Asiyah Yu and Arabandi, Sivaram and Beale, Thomas and Duncan, William D and Hicks, Amanda and Hogan, William R and Jensen, Mark and Koppel, Ross and Mart{\'\i}nez-Costa, Catalina and Nytr{\o}, {\O}ystein and others},
  journal={Standards},
  volume={3},
  number={3},
  pages={316--340},
  year={2023},
  publisher={MDPI}
}

@article{noisy_or,
  title={Learning Bayesian network parameters from small data sets: Application of Noisy-OR gates},
  author={Oni{\'s}ko, Agnieszka and Druzdzel, Marek J and Wasyluk, Hanna},
  journal={International Journal of Approximate Reasoning},
  volume={27},
  number={2},
  pages={165--182},
  year={2001},
  publisher={Elsevier}
}

@article{xgboost_tabular,
  title={Tabular data: Deep learning is not all you need},
  author={Shwartz-Ziv, Ravid and Armon, Amitai},
  journal={Information Fusion},
  volume={81},
  pages={84--90},
  year={2022},
  publisher={Elsevier}
}

@inproceedings{synsum_old,
    title = "SynSUM -- Synthetic Benchmark with Structured and Unstructured Medical Records.",
    author = "Rabaey, Paloma and Arno, Henri and Heytens, Stefan and Demeester, Thomas",
    booktitle = "GenAI4Health, Workshop on Large Language Models and Generative AI for Health at AAAI 2025",
    year = "2025"
}

\newpage
\begin{appendices}

\section{Bayesian network} \label{sec:app_BN}


Table \ref{tab:antibiotics_test_cases} shows the comparison of the probability predicted by the \textit{antibiotics} model (Equation \eqref{eq:antibiotics} in Section \ref{sec:bayesian_network}) with test cases labeled by the expert. Table \ref{tab:days_train_cases} shows the expert-labeled train cases that were used to fit the coefficients in the \textit{\# days} model (Equation \eqref{eq:days_home_no_antibiotics} and \eqref{eq:days_home_antibiotics} in Section \ref{sec:bayesian_network}). Table \ref{tab:days_test_cases} shows the comparison of the probability predicted by the \textit{\# days} model with test cases labeled by the expert.

\begin{table}[!ht]
\centering
\caption{Test cases labeled by the expert on whether to prescribe antibiotics or not (all assume $\text{\textit{policy}} = \text{\textit{low}}$). ``label'' indicates the expert's decision, while ``pred'' indicates the probability predicted by the antibiotics model, according to Equation \eqref{eq:antibiotics}.}
\begin{tabular}{cccccc}
    \toprule
    \multicolumn{4}{c}{\textbf{Symptoms}} & \multicolumn{2}{c}{\textbf{Antibiotics}} \\
    \cmidrule(r){1-4} 
    \cmidrule(r){5-6}
    dysp & cough & pain & fever & label & pred. \\
    \midrule
    no & yes & no & high & no & 0.48 \\
    no & yes & yes & high & yes & 0.64 \\
    yes & yes & no & high & yes & 0.67 \\
    yes & yes & yes & high & yes & 0.80 \\
    yes & no & no & high & yes & 0.51 \\
    no & no & yes & high & yes & 0.48 \\
    no & no & no & high & no & 0.32 \\
    no & no & no & low & no & 0.11 \\
    no & yes & no & low & no & 0.19 \\
    yes & yes & no & low & no & 0.35 \\
    no & yes & yes & low & no & 0.31 \\
    yes & yes & yes & low & yes & 0.51 \\
    yes & yes & yes & none & yes & 0.30 \\
    yes & no & yes & none & no & 0.18 \\
    yes & yes & no & none & no & 0.18 \\
    yes & no & no & none & no & 0.10 \\
    no & yes & no & none & no & 0.09 \\
    no & no & yes & none & no & 0.09 \\
    \bottomrule
    \bottomrule
\end{tabular}
\label{tab:antibiotics_test_cases}
\end{table}

\begin{table}[!h]
\centering
\caption{Train cases labeled by the expert on how many days they expect a patient to stay home (all assuming $\text{\textit{self{\text-}employed}} = \text{\textit{no}}$), with and without prescribing antibiotics. ``label'' indicates the expert's estimation, while ``pred'' indicates the mean number of days $\lambda$ predicted by the \textit{\pr{days at home}} models (Equation \eqref{eq:days_home_no_antibiotics} and \eqref{eq:days_home_antibiotics}).}
\begin{tabular}{ccccccccc}
    \toprule
    \multicolumn{5}{c}{\textbf{Symptoms}} & \multicolumn{4}{c}{\textbf{Days at home}} \\
     &&&&& \multicolumn{2}{c}{\textbf{antibio = no}} & \multicolumn{2}{c}{\textbf{antibio = yes}} \\
    \cmidrule(r){1-5} 
    \cmidrule(r){6-7} \cmidrule(r){8-9}
    dysp & cough & pain & nasal & fever & label & pred. & label & pred. \\
    \midrule
    no & no & no & no & none & 1.5 & 1 & 1 & 1.1\\
    no & yes & no & no & high & 4 & 4.9 & 3.5 & 3.2\\
    no & yes & no & no & low & 2 & 3.2 & 2 & 2.3\\
    no & yes & yes & no & high & 9 & 7.9 & 4 & 4.1\\
    yes & yes & no & no & high & 10 & 9.3 & 5 & 5.3\\
    yes & yes & yes & no & high & 14 & 14.9 & 7 & 6.9\\
    no & yes & yes & no & low & 5 & 5.2 & 3 & 2.9\\
    yes & yes & no & no & low & 6 & 6.1 & 4 & 3.8\\
    yes & yes & yes & no & low & 10 & 9.8 & 5 & 4.9\\
    yes & yes & yes & no & none & 4 & 4.3 & 3.5 & 3.9\\
    no & yes & yes & no & none & 2 & 2.3 & 2 & 2.3 \\
    yes & yes & no & no & none & 3 & 2.7 & 3 & 3\\
    yes & no & yes & no & none & 3 & 3.1 & 3 & 2.5\\
    no & no & no & yes & none & 2 & 1 & 2 & 1.2\\
    
    no & yes & no & yes & high & 4 & 4.7 & 3.5 & 3.2 \\
    no & yes & no & yes & low & 2 & 3.3 & 2 & 2.3\\
    no & yes & yes & yes & high & 9 & 8 & 4 & 4.1\\
    yes & yes & no & yes & high & 10 & 9.4 & 5 & 5.3\\
    yes & yes & yes & yes & high & 14 & 15.1 & 7 & 6.9\\
    no & yes & yes & yes & low & 5 & 5.2 & 3 & 3\\
    yes & yes & no & yes & low & 6 & 6.2 & 4 & 3.8\\
    yes & yes & yes & yes & low & 10 & 9.9 & 5 & 4.9\\
    yes & yes & yes & yes & none & 4 & 4.4 & 3.5 & 3.9\\
    no & yes & yes & yes & none & 2 & 2.3 & 2 & 2.3\\
    yes & yes & no & yes & none & 3 & 2.7 & 3 & 3 \\
    yes & no & yes & yes & none & 3 & 3.1 & 3 & 2.6\\
    \bottomrule
    \bottomrule
\end{tabular}
\label{tab:days_train_cases}
\end{table}

\begin{table}[!h]
\centering
\caption{Test cases (not part of training set) labeled by the expert on how many days they expect a patient to stay home (all assuming $\text{\textit{self{\text-}employed}} = \text{\textit{no}}$), with and without prescribing antibiotics. ``label'' indicates the expert's estimation, while ``pred'' indicates the mean number of days $\lambda$ predicted by the \textit{\pr{days at home}} models (Equation \eqref{eq:days_home_no_antibiotics} and \eqref{eq:days_home_antibiotics}).}
    \begin{tabular}{ccccccccc}
        \toprule
        \multicolumn{5}{c}{\textbf{Symptoms}} & \multicolumn{4}{c}{\textbf{Days at home}} \\
         &&&&& \multicolumn{2}{c}{\textbf{antibio = no}} & \multicolumn{2}{c}{\textbf{antibio = yes}} \\
        \cmidrule(r){1-5} 
        \cmidrule(r){6-7} \cmidrule(r){8-9}
        dysp & cough & pain & nasal & fever & label & pred. & label & pred. \\
        \midrule
        yes & no & no & no & high & 6 & 6.5 & 3.5 & 3.5 \\
        no & no & yes & no & high & 6 & 5.5 & 3 & 2.7 \\
        yes & no & yes & no & high & 12 & 10.5 & 5 & 4.5 \\
        yes & no & no & no & low & 4 & 4.3 & 3 & 2.5 \\
        no & no & yes & no & low & 4 & 3.7 & 3 & 1.9 \\
        yes & no & yes & no & low & 6 & 6.9 & 5 & 3.2 \\
        \bottomrule
    \end{tabular}
\label{tab:days_test_cases}
\end{table}

\section{LLM prompt} 

\subsection{Prompt details} \label{sec:app_prompt_details}

We prompt a large language model to generate a clinical note based on the tabular variables associated with each fictional patient record. The scenario we simulate artificially is as follows. The patient goes to the primary care physician, telling them their symptoms and possible underlying conditions, along with additional context on the severity of these symptoms, when they started, and other details. The physician takes descriptive notes during this consultation, writing down the (recent) history prescribed by the patient. Then, based on the patient's described complaints, they conduct a physical examination, writing down all findings. Both parts together then form the textual description of the patient encounter. \\

We now describe the different parts of the prompt.\\

\textbf{Presence of symptoms} \quad The first block of information in the prompt concerns the symptoms experienced by the patient. We do not list the full set of symptoms exhaustively, since there is a possibility that a patient does not mention a symptom to the clinician, or that the clinician does not find it noteworthy enough to write down. 
We therefore ask our expert to list the probability of each symptom being mentioned in a clinical note, both when the symptom is positive and when it is negative. While we do not expect these arbitrary probabilities to generalize to all settings, it helps to bring some variety and realism in the notes we generate. The probabilities are as follows: 
\begin{itemize}
    \item 
    $\mathcal{P}(\text{\textit{ment}}_{\text{\textit{dysp}}} = \text{\textit{yes}} \mid \text{\textit{dysp}} = \text{\textit{yes}}) = 0.95$,
    $\mathcal{P}(\text{\textit{ment}}_{\text{\textit{dysp}}} = \text{\textit{yes}} \mid \text{\textit{dysp}} = \text{\textit{no}}) = 0.75$
    \item 
    $\mathcal{P}(\text{\textit{ment}}_{\text{\textit{cough}}} = \text{\textit{yes}} \mid \text{\textit{cough}} = \text{\textit{yes}}) = 0.95$,
    $\mathcal{P}(\text{\textit{ment}}_{\text{\textit{cough}}} = \text{\textit{yes}} \mid \text{\textit{cough}} = no) = 0.9$
    \item 
    $\mathcal{P}(\text{\textit{ment}}_{\text{\textit{pain}}} = \text{\textit{yes}} \mid \text{\textit{pain}} = \text{\textit{yes}}) = 0.75$,
    $\mathcal{P}(\text{\textit{ment}}_{\text{\textit{pain}}} = \text{\textit{yes}} \mid \text{\textit{pain}} = \text{\textit{no}}) = 0.3$
    \item 
    $\mathcal{P}(\text{\textit{ment}}_{\text{\textit{fever}}} = \text{\textit{yes}} \mid \text{\textit{fever}} = \text{\textit{high}}) = 0.95$,
    $\mathcal{P}(\text{\textit{ment}}_{\text{\textit{fever}}} = \text{\textit{yes}} \mid \text{\textit{fever}} = \text{\textit{low}}) = 0.7$,
    $\mathcal{P}(\text{\textit{ment}}_{\text{\textit{fever}}} = \text{\textit{yes}} \mid \text{\textit{fever}} = \text{\textit{none}}) = 0.4$
    \item 
    $\mathcal{P}(\text{\textit{ment}}_{\text{\textit{nasal}}} = \text{\textit{yes}} \mid \text{\textit{nasal}} = \text{\textit{yes}}) = 0.95$,
    $\mathcal{P}(\text{\textit{ment}}_{\text{\textit{nasal}}} = \text{\textit{yes}} \mid \text{\textit{nasal}} = \text{\textit{no}}) = 0.1$
\end{itemize}

For each symptom, we sample whether it is to be mentioned in the prompt, conditional on its value, according to the probabilities stated above. As can be seen in Fig. \ref{fig:data_generation}, we explicitly tell the model what symptoms to mention and which to steer clear from. We randomly permute the ordering of the symptoms in each prompt. \\

\begin{table*}[t]
\centering
\caption{Descriptors used in the prompt to describe each symptom when it is present in the patient. Depending on the cause(s) of the symptom (as present in the tabular patient record), we randomly sample from a different set of descriptors.}
\resizebox{\textwidth}{!}{
\begin{tabular}{lll}
    \toprule
    \textbf{Symptom} & \textbf{Cause} & \textbf{Descriptors} \\
    \midrule
    \textbf{dyspnea} & asthma & \makecell[lt]{attack-related, at night, in episodes, wheezing, difficulty breathing in, feeling of suffocation, \\ nighttime stuffiness, provoked by exercise, light, severe, not able to breathe properly, air hunger} \\ 
    & smoking & \makecell[lt]{during exercise, worse in morning, mild} \\
    & COPD & \makecell[lt]{chronic, worse during flare-up, worse when lying down, difficulty sleeping, air hunger} \\
    & hay fever & \makecell[lt]{light, mild, stuffy feeling, all closed up} \\
    & pneumonia & \makecell[lt]{light, mild, severe, no clear cause} \\
    \textbf{cough} & asthma & \makecell[lt]{attack-related, dry} \\
    & smoking & \makecell[lt]{productive, mostly in morning, during exercise, gurgling} \\
    & COPD & \makecell[lt]{phlegm, sputum, gurgling, worse when lying down} \\
    & pneumonia & \makecell[lt]{for over 7 days, light, mild, severe, non-productive at first and later purulent} \\
    & common cold & \makecell[lt]{prickly, irritating, dry, phlegm, sputum, light, mild, severe, constant, day and night} \\
    \textbf{pain} & asthma & \makecell[lt]{tension behind sternum} \\
    & COPD & \makecell[lt]{light, mild} \\
    & cough & \makecell[lt]{muscle pain, burning pain in trachea, burning pain in windpipe, scraping pain in trachea, \\ scraping pain in windpipe} \\
    & pneumonia & \makecell[lt]{light, mild, severe, localized on right side, localized on left side, associated with breathing} \\
    & common cold & \makecell[lt]{burning pain in trachea, burning pain in windpipe, scraping pain in trachea, \\scraping pain in windpipe, light, mild} \\
    \bottomrule
    \bottomrule
\end{tabular}}
\label{tab:symptom_descriptors}
\end{table*}

\textbf{Symptom descriptors}\quad To make the note realistic, the LLM must invent some context regarding the patient's symptoms when writing the history portion of the note. We want this context to indirectly relate to the cause of these symptoms, as they would in a real patient encounter. For example, a cough induced by asthma would likely be momentarily and attack-related, while a cough resulting from pneumonia might be more persistent over the longer term. We therefore ask the expert to write down a list of adjectives or phrases describing each symptom, conditioned on the cause of the symptom. These descriptors can be found in Table \ref{tab:symptom_descriptors}. The list of possible causes for a symptom is simply the list of parents in the Bayesian network. 

For each symptom which is present in the patient and selected to be mentioned in the note, we check the tabular patient record for the possible causes. For example, for the symptom cough, the possible causes are asthma, smoking, COPD, pneumonia and common cold (its parents in the Bayesian network). In the example in Fig. \ref{fig:data_generation}, asthma is the only cause present in the patient. We therefore randomly sample a descriptor from the list of descriptors for cough in the presence of asthma, in this case the adjective ``dry''. This adjective is added in the prompt. If multiple causes are present, we find the strongest cause, and sample from that list. The strongest cause is pneumonia, followed by common cold, followed by all other causes. If neither pneumonia nor common cold is part of the multiple causes, we simply make a bag of all descriptors associated to the causes which are present in the patient, and sample from that bag. In the event that no causes are present, yet a symptom is still observed (which is possible due to the leak probabilities in the Noisy-OR distributions), we do not add a descriptor. Note that while the diagnoses pneumonia and common cold should not be mentioned explicitly, they indirectly and subtly influence the content of the note through the descriptors, adding another realistic dimension to the content of the note. \\

\textbf{Underlying health conditions} \quad While the diagnoses should not be mentioned directly in the note, it is realistic to assume that the note would mention underlying health conditions the patient may have. Since these health conditions are assumed to be known up-front, as they are part of the history of the patient, they may contribute a lot to the interpretation of the symptoms by both the patient themselves and the clinician writing down the note. We therefore add them to the prompt as well, as can be seen in Fig. \ref{fig:data_generation}. We do not force the LLM to explicitly mention these in the note, since it seems feasible that a clinician would not mention them every time. Should there be more than one underlying condition, we mention them all, randomly permuting the order. If there are no underlying health conditions, we simply remove this part of the prompt. \\

\textbf{Additional instructions} \quad We tell the LLM that the note must be structured with a ``History'' portion and a ``Physical examination'' portion. While the ``History'' portion describes the patient's self-reported symptoms and underlying health conditions, which are in large part dictated by the prompt, the ``Physical examination'' portion leaves the LLM with more freedom to imagine additional clinical examinations which were performed on the patient. 
As such, the ``Physical examination'' portion has a lot of potential for adding complexity, clinical terminology and realism to the note. 

We also add some additional instructions to the prompt, asking it not to mention any suspicions of possible diagnoses. We further tell the LLM it can imagine context or details, but no additional symptoms. We noticed that if we left this part out, the LLM would sometimes mention the symptoms we specifically asked to leave out. We ask not to mention patient gender or age, because our initial experiments without this instruction revealed that the LLM often used the same age and gender, which could confound or bias the notes. Finally, we add that the notes may be long (around 5 lines or more), to avoid the LLM being too succinct. \\

\textbf{Model details and parameters} \quad As a large language model, we opted for OpenAI's GPT-4o model, using the version released in May 2024 \citep{GPT4o}. We set the temperature to 1.2 to encourage some more variation in the notes. Before providing the case-specific prompt, we set the following system message: ``You are a general practitioner, and need to summarize the patient encounter in a clinical note. Your notes are detailed and extensive.'' We set the \textit{max\_tokens} parameter to 1000. All other parameters were set to their default values. Generating all 10{,}000 notes and their compact version cost around \$$150$.

\subsection{Additional example prompts} \label{sec:app_example_prompts}

\begin{figure*}[!ht]
\centering
\includegraphics[width=\textwidth]{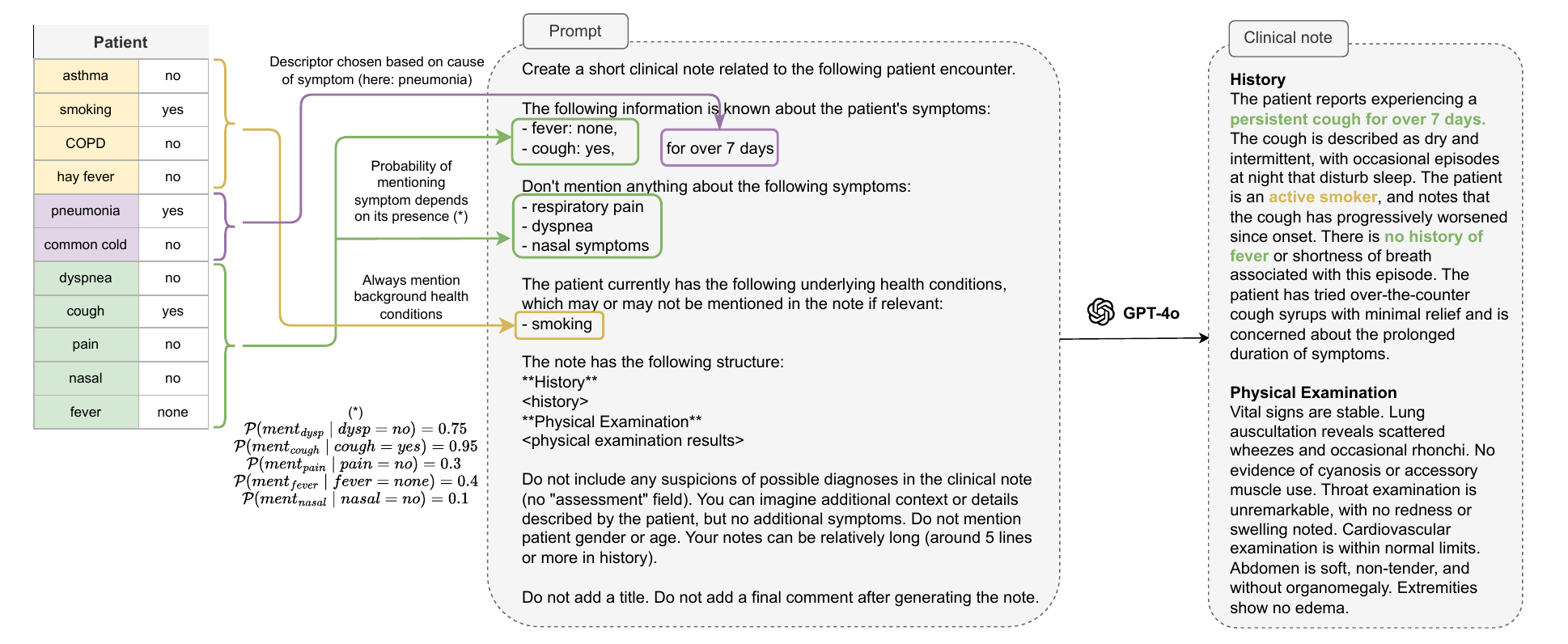}
\caption{Additional example prompt with generated note.}
\label{fig:additional_example_1}
\end{figure*}

\begin{figure*}[!ht]
\centering
\includegraphics[width=\textwidth]{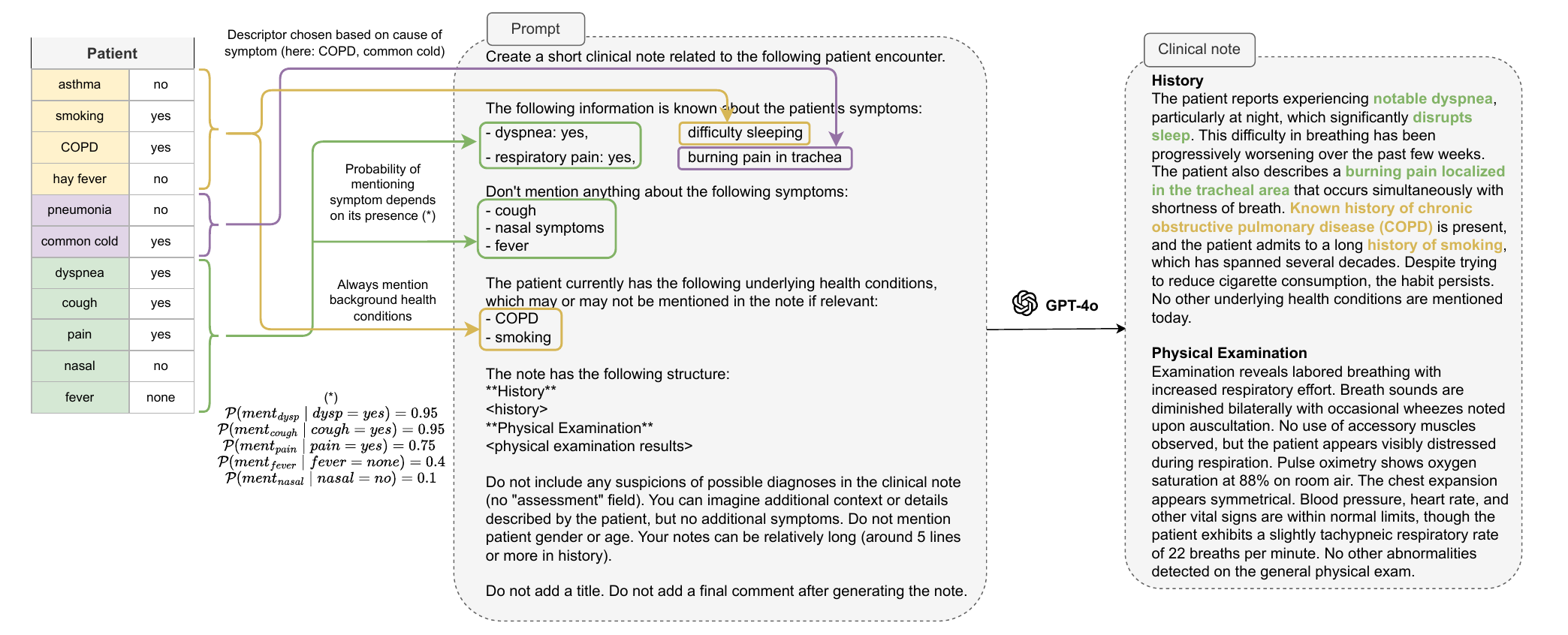}
\caption{Additional example prompt with generated note. Since there are multiple causes for the symptom dyspnea (smoking, COPD) which are present in the patient, the descriptor ``difficulty sleeping'' was chosen randomly out of the bag of descriptors for smoking and COPD from table \ref{tab:symptom_descriptors}. For respiratory pain, the descriptor ``burning pain in the trachea'' was chosen randomly out of the descriptors for common cold. While cough and COPD are also possible causes for respiratory pain, and are present in the patient, common cold overrules the two according to our strategy outlined in Section \ref{sec:prompting_strategy}.}
\label{fig:additional_example_2}
\end{figure*}

Fig. \ref{fig:additional_example_1} and Fig. \ref{fig:additional_example_2} show two additional example prompts.

\subsection{Prompting strategy for special cases} \label{sec:app_prompting_special_cases}

\begin{figure*}[!ht]
\centering
\includegraphics[width=\textwidth]{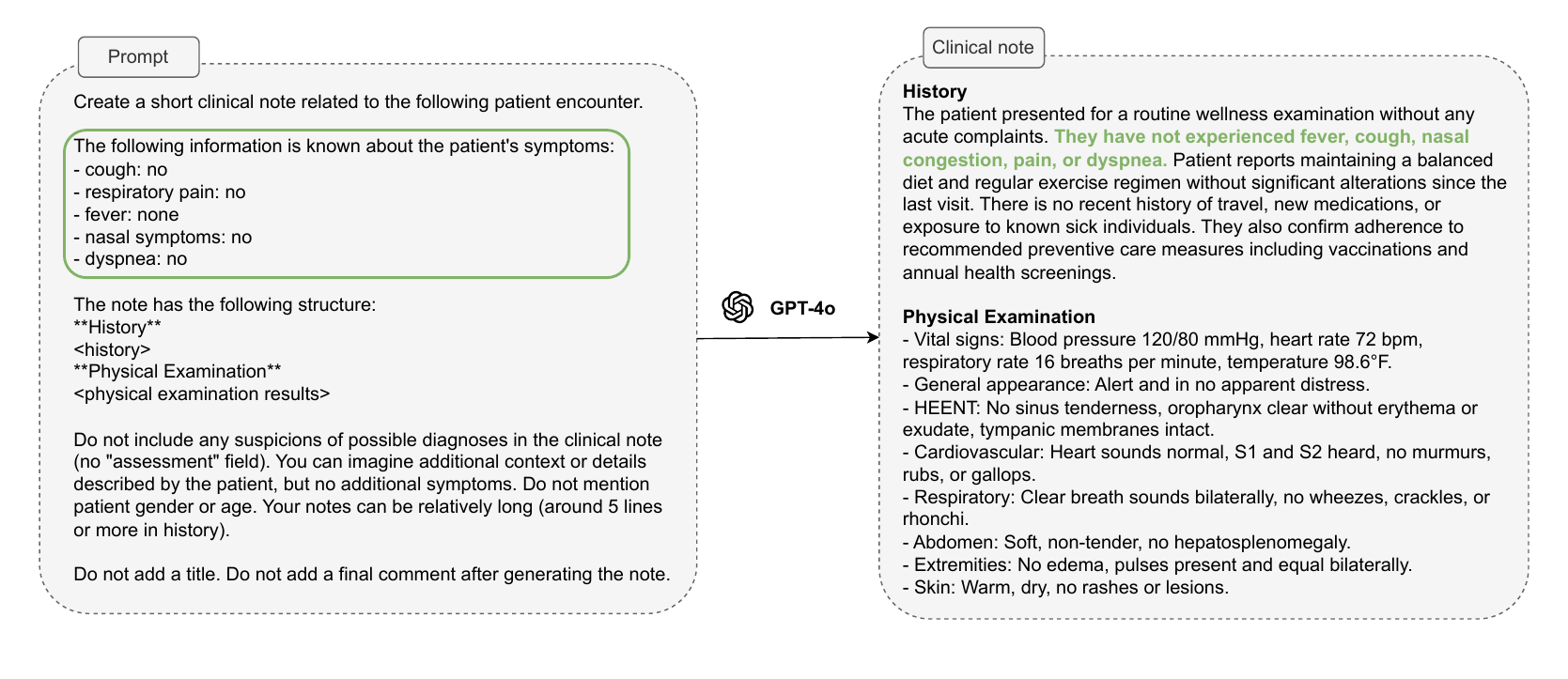}
\caption{Example of what would happen if we simply extended the general prompt to the 3629 cases where the patient does not experience any respiratory symptoms. There would be little variation in these notes, 
as these patients seem to visit the doctor's office for no reason. 
}
\label{fig:bad_example_unrelated}
\end{figure*}

Around one third of all patients in our dataset (3629 out of 10{,}000 patients sampled from the BN) do not experience any respiratory symptoms at all (all symptoms are ``no''). If we used the same prompt as before, this would result in an unrealistic clinical note, since the note would simply list all symptoms the patient does not have, without giving an actual reason for the patient's visit. Furthermore, there would be little variation in these notes. An example of what would happen should we use the original prompt is shown in Fig. \ref{fig:bad_example_unrelated}. 

\begin{figure*}[!ht]
\centering
\includegraphics[width=\textwidth]{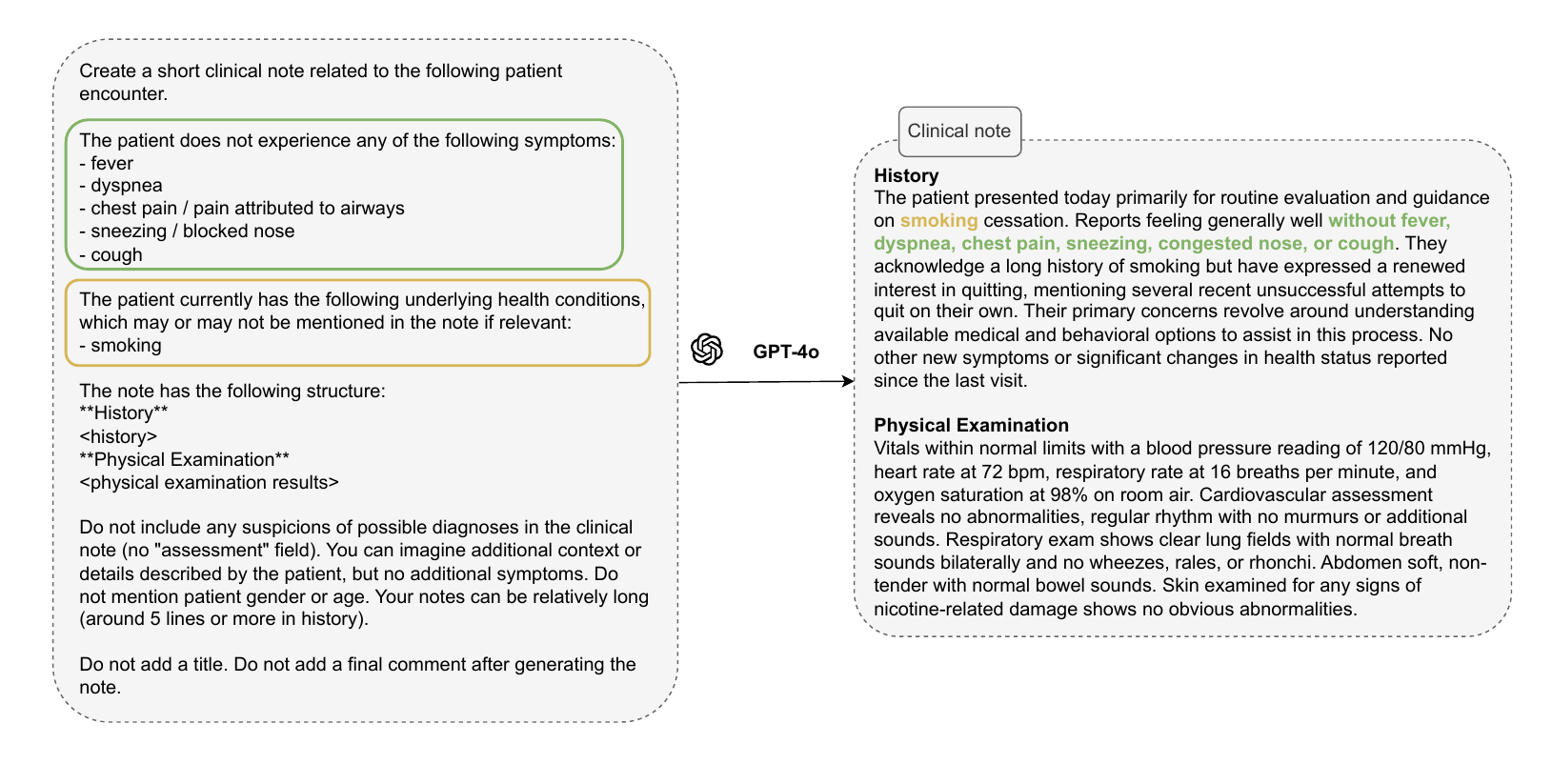}
\caption{Example prompt and generated note for case where patient does not experience any respiratory symptoms, but does have an underlying respiratory health condition (here: smoking).%
}
\label{fig:prompt_unrelated_bg}
\end{figure*}

\begin{figure*}[!ht]
\centering
\includegraphics[width=\textwidth]{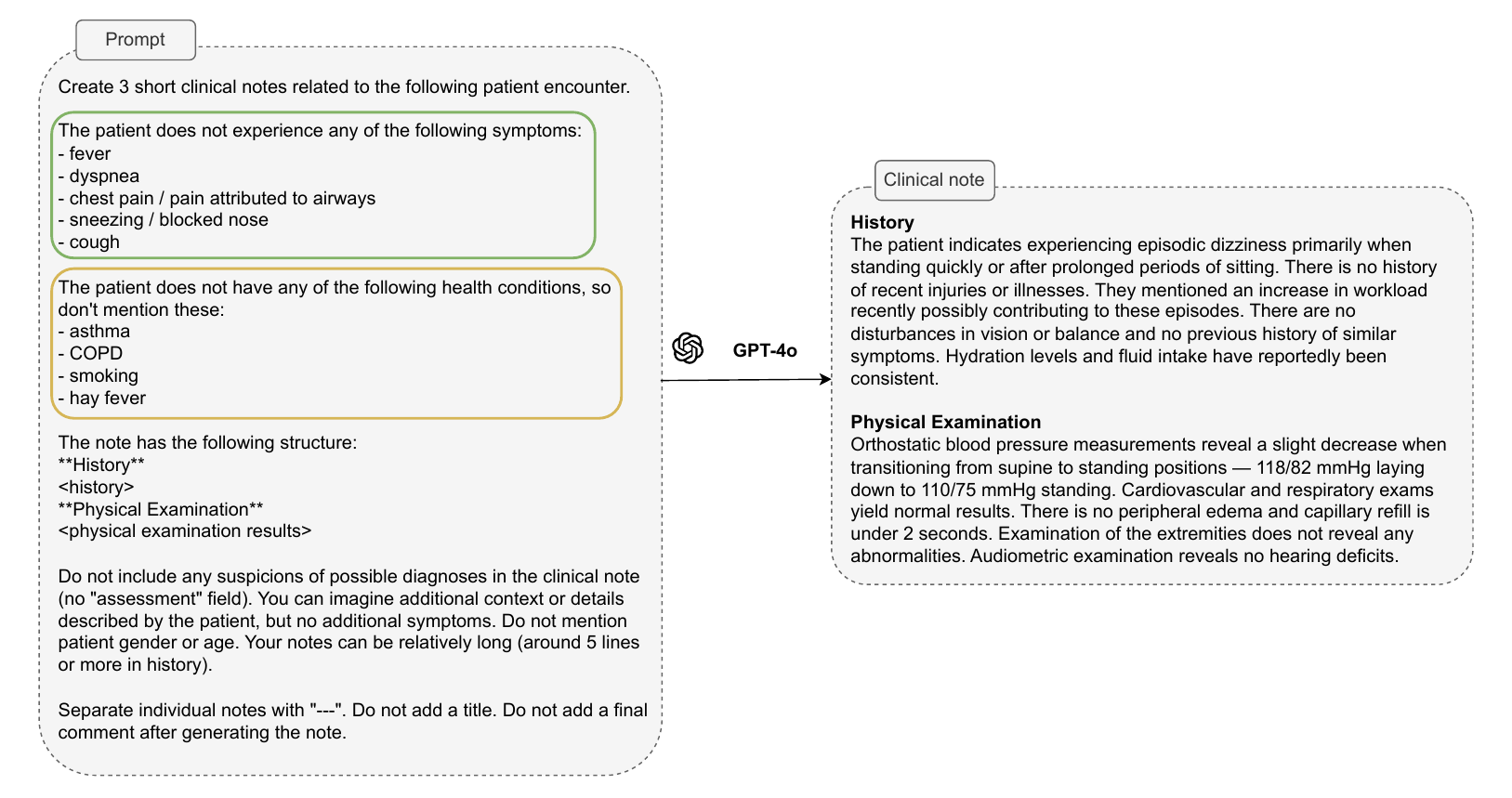}
\caption{Prompt used for cases where patient does not experience any respiratory symptoms or have any underlying respiratory health condition. We show one of the three generated notes.%
}
\label{fig:prompt_unrelated_no_bg}
\end{figure*}

In these cases, it makes more sense to assume that the patient visits the doctor because of a complaint unrelated to the respiratory domain, such as back pain, stomach issues, a skin rash, etc. To generate these special cases, we use a special prompt, telling the LLM the patient does not experience any of the 5 respiratory symptoms. When the patient has at least one underlying health condition (which is the case in 239 out of 3629 special cases), we add this to the prompt in the same way as before, like the example in Fig. \ref{fig:prompt_unrelated_bg}. If not (i.e., for the remaining 3390 out of 3629 special cases), we tell the LLM not to mention any of those health conditions either, as shown in Fig. \ref{fig:prompt_unrelated_no_bg}. The latter prompt asks for three clinical notes at once, encouraging the LLM to be more creative and not repeat the same scenario every time, as well as being a little more cost-effective. This is possible because of the prompt being non-specific to any of these 3390 patients. We then randomly distribute all generated notes to each tabular patient record within this subset. \\

\section{Automated span extraction} \label{app:span_extraction}

To extract spans mentioning each symptom from the normal clinical notes, we send the following instructions to the LLM in the system message: 

\begin{small}
\begin{verbatim}
--- 
I will show you a clinical note containing information on a patient's symptoms. 
For each symptom, I will tell you whether the patient suffers from this symptom 
or not. 

Your task is to extract phrases from the note that mention these symptoms. 
The annotation must have the following JSON structure:  
[ 
   {
      "symptom": one of the symptoms ("dyspnea", "cough", "respiratory pain", 
      "fever" or "nasal symptoms")
      "text": phrase in the text that mentions the symptom and whether it is
      present or absent
   }  
   {
      "symptom": ...
      "text":...
   }
   ...
]

Keep the following instructions in mind:  
- The same symptom may be mentioned multiple times. Include all phrases in 
which a symptom is mentioned. Consider both the "history" portion of the note, 
and the "physical examination" portion of the note.
- Also annotate a symptom if the note mentions that the patient does not
suffer from it. 
- The phrases do not need to be full sentences, but need to be verbatim as 
they appear in the note. You are not allowed to alter any words. If you 
leave out words, use ...
- Order does not matter.
- You will reply only with the JSON itself, and you will not wrap in JSON 
markers.
- You can only extract phrases from the "clinical note", not from any of 
the other text in the prompt. 
- Not all symptoms are necessarily mentioned in the note. Do not include a 
symptom in the JSON if you cannot find any implicit or explicit mention of 
it in the clinical note.
---
\end{verbatim}
\end{small}

We then ask the LLM to extract the spans for each note, using the prompt shown in Fig. \ref{fig:span_extraction_normal}. 

For the compact notes, we tell the LLM to match the spans extracted from the normal note to the text in the compact note. We send the following instructions to the LLM in the system message: 

\begin{small}
\begin{verbatim}
---
I will show you two versions of a clinical note. The first version describes a 
patient's visit to the doctor's office. The second one describes the same visit, 
but in a more compact style (using abbreviations and shortcuts), while preserving 
the overall message. 

I will show you a set of phrases which were extracted from the first version of 
the note. Your task is to map these to phrases in the second version of the note. 

You must reply with the following JSON format.
{
   phrase in version 1 : corresponding phrase extracted in version 2
}

Keep the following instructions in mind: 
- Please extract phrases verbatim. 
- Please use the empty string if you cannot find a phrase with the same meaning.
- The phrases you extract must have the same meaning, you cannot simply copy 
phrases that are in the same spot in the text.
- You will reply only with the JSON itself, and you will not wrap in JSON markers.
---
\end{verbatim}
\end{small}

We then ask the LLM to match the phrases for each pair of notes, using the prompt shown in Fig. \ref{fig:span_extraction_compact}. After the first response, we ask it to check if the extracted phrase has the same meaning, because we noticed that it often selected phrases because they were in the same spot in the text, even if they did not mean the same as the original phrase. We used the following prompt for this: 

\begin{small}
\begin{verbatim}
---
Please check if each extracted phrase has the same meaning as the original phrase. 
If not, substitute it by the empty string (`'). The rest of the JSON must remain 
unchanged.
---
\end{verbatim}
\end{small}

We post-process all phrases extracted from both the normal and compact notes by matching them to the text. If a phrase does not match the text exactly, we ask the LLM to correct itself. For this, we use the following prompt: 

\begin{small}
\begin{verbatim}
---
I will show you a clinical note, together with one or more phrases that were 
extracted from it. However, some mistakes were made in extracting these 
phrases. You must correct them.

---

The following is a clinical note:
< note >

The following phrases were extracted from this note. However, they do not 
exactly match the text:
- < phrase 1 > 
- < phrase 2 > 
- ...

Please correct the phrases so they map exactly to a phrase in the text. 
You must reply with the following JSON format: 
{
   original phrase: corrected phrase
}

You will reply only with the JSON itself, and you will not wrap in JSON markers.
---
\end{verbatim}
\end{small}

For all these prompts, we use OpenAI's GPT-4o model. We set the temperature to 0.2 to encourage correctness over creativity. The full span annotation process cost around \$80. For a full walk through of the span extraction process, we refer to the tutorial notebook on our Github repository: \url{https://github.com/prabaey/SimSUM/blob/main/src/span_annotations.ipynb}.

\section{Expert evaluation} \label{sec:app_evaluation}

We picked a random subset of $30$ generated notes and show them to $5$ general practitioners. All evaluators got to see the same $30$ notes, together with the prompts that were used to generate them, in a random order that differed between evaluators. All evaluators received detailed instructions on what was expected of them in the form of a PDF\footnote{\url{https://github.com/prabaey/SimSUM/blob/main/eval/Instructions_clinical_evaluation.pdf}}, which was orally explained by the authors. The authors then sat together with each evaluator separately to complete three example notes (different from the $30$ notes that were to be evaluated). Afterwards, the evaluators were asked to rate the notes in their own time, without the authors' involvement. Since it is infeasible to evaluate the whole dataset of $10{,}000$ notes, we opted for a small subset of $30$ notes, each going through a relatively extensive evaluation process that considered various measures of quality (evaluators took around $5-10$ minutes to evaluate each note). We decided to show all $5$ evaluators the same set of $30$ notes, to get a broader range of expert opinions in the evaluation of each note. This also allowed for the calculation of inter-annotator agreement. 

We now provide further details on each dimension along which we evaluated the notes and the specific meaning assigned to each rating. The aspects of consistency, realism and clinical accuracy are only evaluated based on the normal note, while the quality of the compact note is evaluated using the last two dimensions (content and readability).

\subsection{Evaluation dimensions}

\subsubsection{Consistency} 

We subdivided the prompt into four different sections, and asked the evaluators to assign penalties for each section. A penalty was assigned if the requested information in that section was incorrectly mentioned in the note (e.g.~a particular symptom was said to be present in the patient, when the prompt particularly requested the symptom to be absent), or if the requested information was absent from the note (e.g.~a symptom descriptor is not mentioned in the text). The four parts of the prompt were as follows: (i) the symptoms to mention (can be present or absent), (ii) the symptoms not to mention, (iii) the underlying health conditions, and (iv) the additional instructions. 

As explained in Section \ref{sec:prompting_strategy}, around one third of the notes were generated using a second type of prompt, where the LLM is told that the patient does not suffer from any respiratory symptoms or underlying health conditions. Following the prompt in Fig. \ref{fig:prompt_unrelated_no_bg}, there are three parts of the prompt which can be violated (leading to penalties): (i) the respiratory symptoms, which the patient does not have, (ii) the underlying conditions, which the patient does not have, and (iii) the additional instructions. In our random set of $30$ notes, $20$ notes belonged to the first type, and $10$ to the second type. 

Once penalties were assigned, we summed them into a total number of penalties, and converted these into scores from $1$ to $5$. Notes with no penalties get a perfect score of $5$, one penalty corresponds to $4$, two penalties to $3$, three penalties to $2$ and more than three penalties to $1$. 

\subsubsection{Realism} 

The LLM is allowed to invent context and details in light of the information it receives in the prompt, but this must be realistic and relevant to the symptoms experienced by the patient. While some clinical facts might not seem technically incorrect, one might not expect to see them in the note, or it might be unlikely that they would be written down by a real physician. For example, if a patient has a runny nose and no other complaints, most clinicians would not check for abnormalities in lung capacity. Another example is asking whether the patient has recently traveled to an exotic destination because they have a cough.

We ask to score realism of the ``history'' section using the ratings below. The evaluators are specifically instructed to take into account the information mentioned in the prompt.
\begin{description}
    \item \textbf{5} -- All pieces of additional context and details (i.e. outside of the symptoms and background provided in the prompt) are realistic and seem like they belong in the note. 
    \item \textbf{4} -- There are one or two pieces of additional context or details that I would not have mentioned as a physician, or that do not seem relevant (even though they do seem like they belong). 
    \item \textbf{3} -- There are one or two pieces of additional context or details that do not seem like they belong in the note, or do not seem relevant, given the symptoms and background provided in the prompt. 
    \item \textbf{2} -- There are multiple pieces of additional context or details that do not seem like they belong in the note, or do not seem relevant, given the symptoms and background provided in the prompt. 
    \item \textbf{1} -- (Almost) all of the additional context is nonsensical given the symptoms and background provided in the prompt. 
\end{description}

We ask to score realism of the ``physical examination'' section using the ratings below. The evaluators are specifically instructed to take into account the information mentioned in the ``history'' section of the note. 
\begin{description}
    \item \textbf{5} -- All elements in the physical examination are things I would check, given the history and symptoms of the patient, and no important elements are missing.  
    \item \textbf{4} -- There are one or two elements in the physical examination that I probably would not have checked, given the history and symptoms of the patient, but I could see it happen. Some minor elements might be missing, but nothing major. 
    \item \textbf{3} -- There are one or two elements in the physical examination that I would not have checked, or some important elements are missing, given the history and symptoms of the patient.
    \item \textbf{2} -- There are multiple elements in the physical examination that make no sense given the history and symptoms of the patient, or many important elements are missing. 
    \item \textbf{1} -- The physical examination portion of the note seems totally unrealistic. 
\end{description}

\subsubsection{Clinical accuracy}

While the previous section talks about evaluating the realism of the presence of all examinations described in the ``physical examination'' section, here we talk about evaluating the clinical accuracy of these findings. 
Clinical inaccuracies may depend on the context, like physical findings which are not congruent with the history and symptoms of the patient. For example, if the ``history'' portion mentions that the patient has a no fever, then this should not be contradicted in the ``physical examination'' portion with a temperature of 39°C. Clinical inaccuracies may also stand alone. For example, a blood pressure reading of 20/10 mm Hg is impossible to encounter in any patient.

We ask the evaluators to score clinical accuracy of the findings that are mentioned in the ``physical examination'' portion of the note, using the ratings below.  
\begin{description}
    \item \textbf{5} -- There are no mistakes, all reported clinical information is plausible in light of the patient's symptoms and history.  
    \item \textbf{4} -- There are one or two minor mistakes, or some details seem less plausible in light of the patient's symptoms and history, while the overall picture painted by the note is still correct. 
    \item \textbf{3} -- There are more than two minor mistakes, or multiple details which seem implausible in light of the patient's symptoms and history, but no major inaccuracies. 
    \item \textbf{2} -- There is a major mistake (on top of possibly some minor ones), or many details seem implausible given the patient's symptoms and history. 
    \item \textbf{1} -- There are multiple major mistakes and many details seem totally implausible given the patient's symptoms and history. 
\end{description}

\subsubsection{Quality of compact version}

While all the previous evaluations concerned the original note, here we evaluate the quality of the compact version of the note. 

The \textbf{content} of the compact version should convey the same information as the original text, albeit in a shorter format. This is evaluated jointly for ``history'' and ``physical examination'' using the scoring system below. 
\begin{description}
    \item \textbf{5} -- The compact version conveys the exact same information as the original text. 
    \item \textbf{4} -- The compact version conveys all key points of the original text, leaving out some details here and there. 
    \item \textbf{3} -- The compact version conveys some of the key points of the original text, but misses some as well. 
    \item \textbf{2} -- The compact version conveys some of the same information as the original text, but misses many key points. 
    \item \textbf{1} -- The compact version does not convey the same information as the original text, leaving out almost all key points. 
\end{description}

While we purposefully want these notes to be harder to read and understand for both humans and machines, mimicking the complexity of some real clinical notes, the use of abbreviations should not be excessive. We evaluate \textbf{readability} jointly for ``history'' and ``physical examination'', using the scoring system below.
\begin{description}
    \item \textbf{5} -- The compact version seems understandable without seeing the original. 
    \item \textbf{4} -- The compact version seems mostly understandable without seeing the original, though there are some abbreviations that I would not immediately understand.
    \item \textbf{3} -- Some parts of the compact version seem understandable without seeing the original, but other parts are not. There are some abbreviations that seem far-fetched or are used incorrectly (i.e. these are known to refer to other clinical terms than the way they are used in the text).
    \item \textbf{2} -- Many parts of the compact version would not be understandable without seeing the original. Many abbreviations seem far-fetched or are used incorrectly (i.e. these are known to refer to other clinical terms than the way they are used in the text).
    \item \textbf{1} -- The compact version is impossible to understand without seeing the original. 
\end{description}

\subsection{Results} \label{sec:app_evaluation_results}

\subsubsection{Extended results for consistency}

We list the total number of penalties assigned by all evaluators to each of the prompt sections. For the 20 notes belonging to the first prompt type, 41 penalties were assigned in total for section (iv) of the prompt, which are the additional instructions. Sections (i), (ii), and (iii), which describe the symptoms and underlying health conditions of the patients, received no penalties at all. For the 10 notes belonging to the second prompt type, 1 penalty was assigned for part (i), 2 penalties for part (ii) and another 2 penalties for part (iii).

\subsubsection{\rebut{Inter-Rater Reliability}}

\rebut{Krippendorff’s alpha is a statistical measure of inter-rater reliability that quantifies the extent to which observed agreement among raters exceeds what would be expected by chance. Usually, an alpha of 0.667 is cited as the desired level in order to draw tentative conclusions \citep{krippendorff}. We calculate Krippendorff's alpha with ordinal distance measure between the 5 evaluators in our study, for each of the measured criteria:}
\begin{itemize}
    \item Consistency: 0.44
    \item Realism (history): 0.25
    \item Realism (physical examination): 0.32
    \item Clinical accuracy: 0.21
    \item Content of compact version: -0.02
    \item Readability of compact version: 0.36
\end{itemize}

With an alpha of 1 indicating perfect agreement, and \rebut{0 indicating agreement at the level of chance}, we can say that there is some agreement between the 5 evaluators. 
\rebut{While we do not reach the desired level of 0.667, in our case we must consider that} due to the high scores assigned by all evaluators (mostly 4 and 5), a deviation from 5 can more easily be attributed to random chance rather than to a shared opinion of the evaluators. The lower standard deviations in Table \ref{tab:evaluation_results} show that the evaluators consistently assign high scores to all aspects of the evaluation, even if they \rebut{do not} agree on which particular notes deviate from these higher scores. 

\section{Symptom predictor baselines}

\begin{figure*}[t]
\centering
\includegraphics[width=0.8\textwidth]{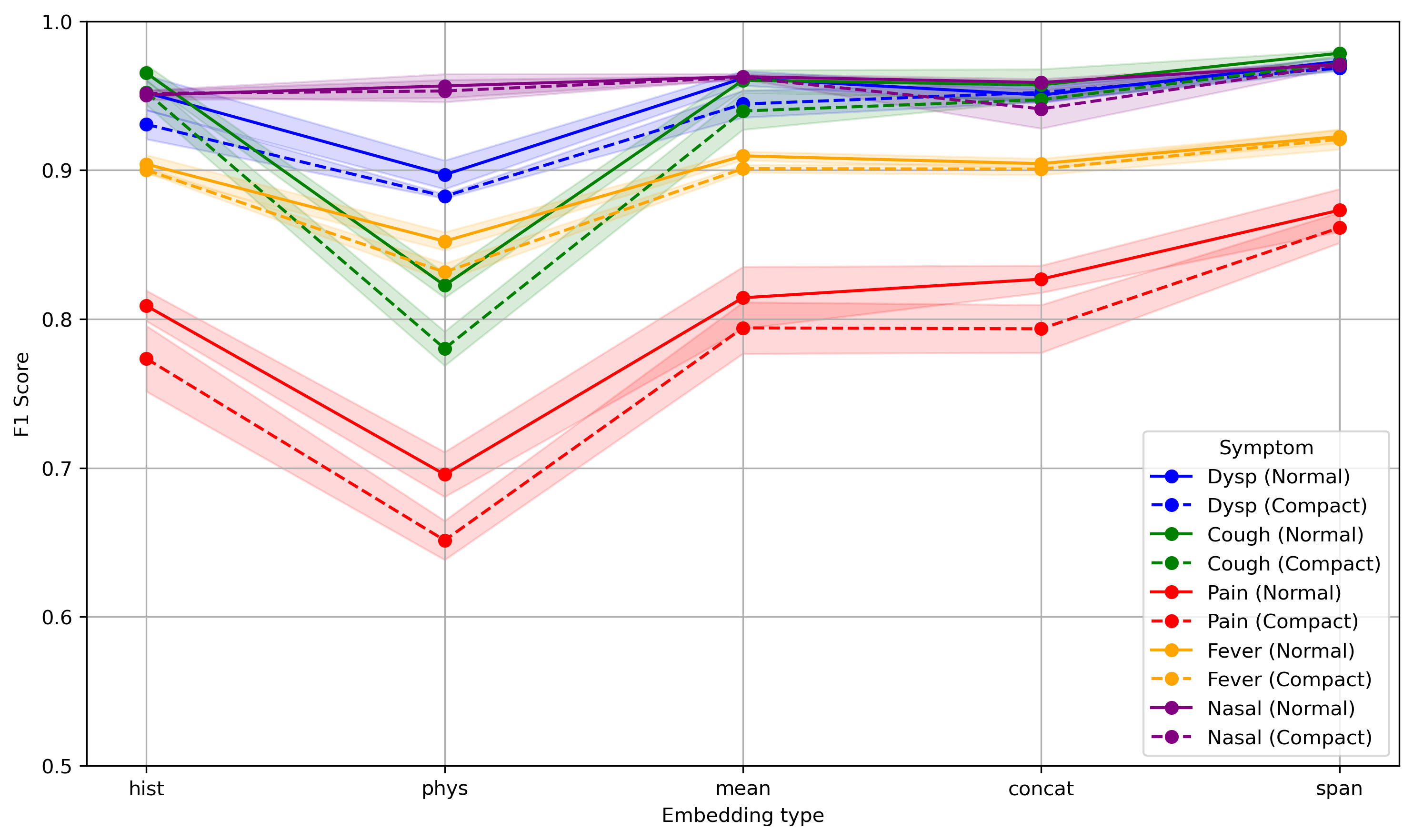}
\caption{Results for the \textbf{neural-text} baseline for the five different embedding types. We show F1-scores over the test set for both the normal and compact versions of the notes, averaging over 5 seeds. Figure best viewed in color. For numerical results, see Table \ref{tab:text_classifier_results} in Section \ref{sec:CIE_baselines}.}
\label{fig:text_emb_results}
\end{figure*}

\subsection{BN-tab} \label{app:BN-tab}

We learn a Bayesian network over the training data, providing the structure over all variables as in Fig. \ref{fig:data_generation}. For the variables \textit{asthma}, \textit{smoking}, \textit{hay\:fever}, \textit{COPD}, \textit{season}, \textit{pneumonia}, \textit{common\:cold}, \textit{fever} and \textit{self{\text-}employed}, we learn the conditional probability tables (CPTs) from the training data using maximum likelihood estimation (which comes down to counting co-occurrences of child and parent values for each entry in the CPT). We use the \code{pgmpy} library \citep{pgmpy} with a K2 prior as a smoothing strategy to initialize empty CPTs. 

As support for learning Noisy-OR distributions is not provided in \code{pgmpy}, we learn these parameters with a custom training loop. We formulate the likelihood as in Equation \eqref{eq:noisy_OR} in Section \ref{sec:bayesian_network}, and learn the parameters $p_i$ in Equations \eqref{eq:noisy_OR_dysp} through \eqref{eq:noisy_OR_nasal} for the variables \textit{dyspnea}, \textit{cough}, \textit{pain} and \textit{nasal} through maximum likelihood estimation by iterating over the train set for 10 epochs, using an Adam optimizer with a batch size of 50, a learning rate of 0.01 and random initialization of each parameter. To integrate the learned Noisy-OR distributions in the Bayesian network, we turn them into fully specified CPTs. To obtain these, we simply evaluate Equation \eqref{eq:noisy_OR} in Section \ref{sec:bayesian_network} for all possible combinations of child and parent values. While this results in large and inefficient CPTs, the automated inference engine built into the \code{pgmpy} library does not support Noisy-OR distributions directly. Note that both versions of the conditional distribution are equivalent, so we do not incur a loss in precision. 

Similarly, the coefficients in the logistic regression model for \textit{antibiotics} and the Poisson regression model for \textit{\pr{days at home}} are learned using maximum likelihood estimation over the training set of $8{,}000$ examples. The likelihood is expressed as in Equation \eqref{eq:antibiotics}, \eqref{eq:days_home_no_antibiotics} and \eqref{eq:days_home_antibiotics} in Section \ref{sec:bayesian_network} respectively, with learnable parameters in place of each coefficient. We iterate over the train set for 15 epochs, again using an Adam optimizer with a batch size of 50, a learning rate of 0.01 and random initialization of each parameter. Finally, we turn the logistic regression and Poisson regression models into CPTs by evaluating Equation \eqref{eq:antibiotics}, \eqref{eq:days_home_no_antibiotics} and \eqref{eq:days_home_antibiotics} for all combinations of parent and child values. For the variable \textit{\pr{days at home}}, we needed to turn each discrete number of days into a category, because \code{pgmpy} only provides automated inference for Bayesian networks consisting of exclusively categorical variables. This results in a large CPT containing one row per possible number of days, which range from 0 to 15 in our training dataset, and one column for each combination of the 7 parent variables. To allow for a possible larger maximum number of days in the test set, we create a category $\geq\text{\textit{15 days}}$, which is defined as one minus the summed probability of all other days. 

Once we have learned all parameters in the joint distribution, we can evaluate the Bayesian network’s ability to predict each of the symptoms. For each evidence setting (as defined in the main text), we apply variable elimination with each of the symptoms as a target variable. Looking at the causal structure in Fig. \ref{fig:data_generation}, we note that the model never has to marginalize over the many rows in the learned \textit{\pr{days at home}} CPT, since it is never a target variable. This makes automated inference feasible in our case. 

\subsection{XGBoost-tab} \label{app:XGBoost-tab}

We use the \code{xgboost} library in combination with \code{sklearn}. We train separate classifiers per symptom, one for each setting, which means we train 15 classifiers total. We tune the hyperparameters separately for each classifier, using 5-fold cross validation with F1 as a scoring metric (macro-F1 for \textit{fever}). 

The classifiers for the symptoms \textit{dysp}, \textit{cough}, \textit{pain} and \textit{nasal} use a binary logistic objective and logloss as an evaluation metric within the XGBoost training procedure, while the classifiers for the symptom \textit{fever} use the multi-softmax objective with multiclass logloss as an evaluation metric. The \textit{scale\_pos\_weight} parameter is set to the ratio of negative over positive samples for the binary classifiers. 
For the \textit{fever} classifier, we address class imbalance by setting \textit{class\_weight = balanced}, which ensures that samples from less frequent classes (in our case low and high fever) receive higher weight in the loss calculation. We use grid search to find the best hyperparameter configuration, where the following sets of options are explored: 
\setlist{nolistsep}
\begin{itemize}[noitemsep]
    \item \textit{n\_estimators}: $\{50, 100, 200\}$
    \item \textit{max\_depth}: $\{2, 3, 4, 5\}$
    \item \textit{learning\_rate}: $\{0.01, 0.1, 0.2\}$
    \item \textit{subsample}: $\{0.8, 1\}$
    \item \textit{colsample\_bytree}: $\{0.8, 1\}$
    \item \textit{gamma}: $\{0, 0.1, 0.3\}$
    \item \textit{min\_child\_weight}: $\{1, 5, 10\}$
\end{itemize}

\subsection{Neural-text} \label{app:neural-text}

The pretrained BioLORD encoder \citep{biolord} was obtained through the \code{huggingface} library. The encoder outputs 768-dimensional sentence embeddings. Since the full text did not fit into the context window, we embedded each sentence separately, and then combined them using our strategies outlined in the main text. To split the text into sentences, we used the \code{nltk} package. The settings \textit{hist}, \textit{phys}, \textit{mean} and \textit{span} all result in a text embedding of 768 dimensions, while the setting \textit{concat} results in a text embedding of 2*768 dimensions. 

These embeddings are fed into a linear layer with 256 neurons, followed by a ReLU activation. The hidden state is then transformed into a single output neuron, followed by a Sigmoid activation. For the classifiers that predict \textit{fever}, three output neurons followed by a Softmax activation are used instead, one for each class. While the embeddings remain fixed, we learn the parameters in the hidden and output layers using cross-entropy as a loss function over the training set. We train a separate classifier for each symptom, setting and difficulty of the text (normal vs. compact). For the binary symptoms, we train for $15$ epochs using the Adam optimizer with a batch size of $100$, a learning rate of $0.001$ and \textit{weight\_decay} set to $1\mathrm{e}-5$. The classifier for $fever$ tended to collapse more easily, which is why we train it for $30$ epochs with a lower learning rate of $5\mathrm{e}-4$ instead. These hyperparameters were obtained using a mix of manual tuning and grid search with 5-fold cross validation over the training set. 

Results for the different embedding types are reported numerically in Table \ref{tab:text_classifier_results} in Section \ref{sec:CIE_baselines}, and shown visually in Fig. \ref{fig:text_emb_results}. 

\subsection{Neural-text-tab} \label{app:neural-text-tab}

For each evidence setting, we select the relevant set of tabular features and transform them into a numerical representation. We use a one-hot encoding for the categorical (binary or multiclass) features, and normalize the \textit{\pr{days at home}} feature using the StandardScaler from \code{sklearn}. This tabular feature representation is then concatenated with the text representation we obtained in the previous baseline. Both are fed into the same architecture described in Section \ref{app:neural-text}, adapting the dimension of the input layer accordingly. For example, for the \textit{dyspnea} classifier in the evidence setting \textit{all}, the input dimension becomes $768 + 17$. We use the same hyperparameters as in the \textbf{neural-text} baseline to ensure a fair comparison. All other training details remain the same. 

\end{appendices}

\end{document}